%% file: aaai2027.tex
\documentclass[letterpaper]{article} % DO NOT CHANGE THIS
\usepackage[preprint]{aaai2027}  % DO NOT CHANGE THIS
\usepackage[hyphens]{url}  % DO NOT CHANGE THIS
\usepackage{graphicx} % DO NOT CHANGE THIS
\usepackage{natbib}  % DO NOT CHANGE THIS AND DO NOT ADD ANY OPTIONS TO IT
\usepackage{caption} % DO NOT CHANGE THIS AND DO NOT ADD ANY OPTIONS TO IT
\usepackage{latexsym}

\usepackage[T1]{fontenc}
\usepackage[utf8]{inputenc}

\usepackage{microtype}

\usepackage{inconsolata}
\usepackage{makecell}
\usepackage{booktabs}
\usepackage{latexsym}
\usepackage{graphicx}
\usepackage{amsthm,amsmath,amssymb,bm}
\usepackage{mathrsfs}
\usepackage{booktabs}
\usepackage{algorithm}
\usepackage{algpseudocode}
\usepackage{comment}
\usepackage{multirow}
\usepackage{bbding}
\usepackage{array}
\usepackage{tcolorbox}
\usepackage{colortbl}
\usepackage{xcolor}
\usepackage{xspace}
\usepackage{enumitem}
\usepackage{threeparttable}
\usepackage{pifont}
\usepackage{makecell}

\newcommand{\blanksymbolfootnote}[1]{%
  \renewcommand{\thefootnote}{}% Remove footnote number
  \footnote{#1}%
  \setcounter{footnote}{0} % Reset footnote counter
  \renewcommand{\thefootnote}{\arabic{footnote}}% Restore footnote number
}

\title{MTAC-IFBench: Benchmarking Instruction-Following in Multi-Turn Agentic Coding}
\author{Bosi Wen$^{1,\dagger, *}$ \quad Cunxiang Wang$^{2,*, \ddagger}$\quad Jiayi Gui$^{2}$\quad Haoke Zhang$^{2}$\quad Yilin Niu$^{2}$\quad  \\
\textbf{Pei Ke$^{3}$\quad Dayong Yang$^{2}$\quad Hongning Wang$^{1}$ \quad Minlie Huang$^{1,\ddagger}$}}

\affiliations{
    \textsuperscript{\rm 1}The Conversational Artificial Intelligence (CoAI) Group, Tsinghua University\\
    \textsuperscript{\rm 2}Zhipu AI \quad \textsuperscript{\rm 3}University of Electronic Science and Technology of China \\
    \texttt{wbs23@mails.tsinghua.edu.cn}, \texttt{aihuang@tsinghua.edu.cn} \\
}

\begin{document}

\maketitle

\blanksymbolfootnote{$^\dagger$Work done when this author interned at Zhipu AI.}
\blanksymbolfootnote{$^*$Equal contribution}
\blanksymbolfootnote{$^\ddagger$Corresponding author}

\begin{abstract}
\input{sections/abstruct}
\end{abstract}

\begin{links}
    \link{Code}{https://github.com/abelperry/AgentProbe}
    \link{Datasets}{https://huggingface.co/datasets/thu-coai/MTAC-IFBench}
\end{links}

\section{Introduction}
\input{sections/introduction}

\section{Related Work}
\input{sections/related_work}

\section{MTAC-IFBench}
\input{sections/methods}

\section{Experiments}
\input{sections/experiments}

\section{Conclusion}
\input{sections/conclusion}

\bibliography{custom}
\appendix
\input{sections/appendix}

% \clearpage
% \input{ReproducibilityChecklist}
% \clearpage


\end{document}

%% file: sections/abstruct.tex
Recently, the rapid development of large language models (LLMs) has reshaped software engineering by enabling autonomous code agents that plan, execute, and utilize external tools iteratively to tackle complex tasks. 
Beyond achieving functional correctness, these agents must faithfully follow process instructions and constraints throughout the development lifecycle. 
However, existing benchmarks typically focus on final functional correctness or confine instruction-following evaluation to single-turn, general chat or simple code generation scenarios, leaving instruction-following in multi-turn agentic coding underexplored. 
To bridge this gap, we propose MTAC-IFBench, a comprehensive benchmark for this critical capability. 
It features multi-turn progressive software development instructions with diverse constraints spanning 6 primary and 18 secondary categories.
With an average of 7.04 turns and 91.33 constraints per instance, it poses a rigorous challenge to current LLMs. 
To make the evaluation reliable, we construct a checklist for each constraint and functional requirement, and integrate verification scripts and judge agents to verify each checklist item.
MTAC-IFBench identifies significant deficiencies in existing code agents in multi-turn instruction-following, with their performance degrading rapidly as the interaction session grows longer.

%% file: sections/introduction.tex
Recently, the rapid development of large language models (LLMs) has fundamentally reshaped software development \cite{sapkota2025vibe}, driving a paradigm shift from conventional human-centric workflows to autonomous agent-based systems \cite{wang2025ai}. 
Agentic coding frameworks, such as Claude Code \cite{claude} and Kilo \cite{kilo}, exemplify this transition by introducing LLM-based agents that autonomously plan, execute, and interact with external tools to iteratively tackle complex tasks \cite{gong2025language}.
During this process, they are required not only to generate functionally correct code, but also to strictly comply with various instructions and constraints throughout the development lifecycle \cite{ding2026octobench}. 
Accurate instruction-following is a fundamental prerequisite for the reliability and practical deployment of code agents in real-world production environments \cite{qi2025agentif}. 

\begin{figure}[t]
\scriptsize
    \centering
    \includegraphics[width=1.0\linewidth]{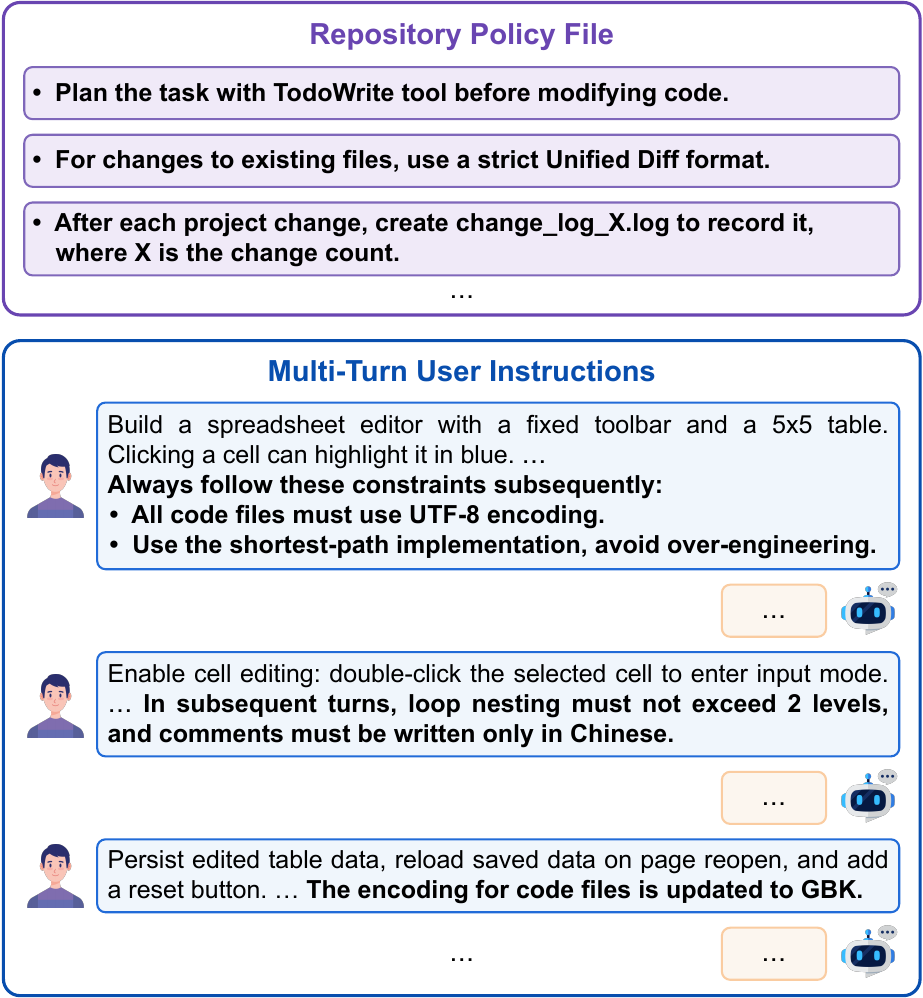}
    \caption{An example from MTAC-IFBench, containing a repository policy file and multi-turn user instructions. Constraints within each instruction are \textbf{bold}, demonstrating dynamic updates such as the addition of new constraints and the replacement of existing ones. }
    \label{fig:example}
    % \vspace{-4mm}
\end{figure}

The agentic coding scenario introduces two major challenges for instruction-following. 
First, constraints in this scenario are highly heterogeneous \cite{ding2026octobench}, encompassing not only generated responses and code, but also environment interactions and agent workflows.
Second, complex software development tasks often involve multi-turn interaction \cite{wu2025frontalk, sweinteract}. 
During the agent's implementation process, users iteratively introduce new requirements or modify existing ones. 
This dynamic necessitates persistent constraint compliance across multiple turns without suffering from context degradation.

\input{tables/comparison_wo_color}

However, existing benchmarks fail to fully capture this reality.
On the one hand, current agentic coding benchmarks, such as SWE-Bench \cite{jimenez2024swe} and Terminal-Bench \cite{merrill2026terminal},  primarily emphasize functional correctness while overlooking compliance with process instructions and constraints.
On the other hand, current instruction-following benchmarks mainly focus on general chat and single-file code generation \cite{zhou2023instruction, wen2024benchmarking, yan2025codeif, wang2025codeif}, leaving a substantial gap in agentic scenarios. 
Although some benchmarks \cite{qi2025agentif, chen2026agentif, ding2026octobench} attempt to expand instruction-following to agentic settings, they remain confined to single-turn interactions and limited constraint types, thereby hindering comprehensive evaluation.

To address these gaps, we introduce MTAC\footnote{MTAC stands for Multi-Turn Agentic Coding.}-IFBench, a benchmark designed to comprehensively evaluate the instruction-following abilities of code agents in multi-turn software development scenarios. 
As shown in Figure \ref{fig:example}, each instance of MTAC-IFBench contains progressive, multi-turn software development instructions that integrate both functional requirements and specific constraints. 
Repository policy files (e.g., Claude.md) are also incorporated to impose global requirements governing the entire development workflow.
% Building upon ZfrontBench \cite{zeng2026glm} and CC-Bench\footnote{https://huggingface.co/datasets/zai-org/CC-Bench-trajectories}, two datasets of single-turn software development instructions,
After collecting single-turn software development instructions from diverse sources, 
we employ LLM-based instruction revision and expansion to synthesize multi-turn task instructions. 
Subsequently, a turn-wise, iterative integration and validation pipeline is employed to integrate constraints for each turn, which is guided by a comprehensive constraint taxonomy of agentic coding encompassing 6 primary and 18 secondary categories. 
Finally, rigorous manual inspections are conducted to guarantee high data quality. 
In total, MTAC-IFBench comprises 100 instances, with an average of 7.04 turns per instance and 12.97 constraints per turn.
For evaluation, we construct checklists for constraint compliance at each turn and functional correctness of the final project separately, utilizing either verification scripts or LLM-based judge agents to verify each checklist item according to its type.
This hybrid approach guarantees reliable assessments for complex requirements.

We conduct a comprehensive evaluation of 11 advanced LLM-based code agents on MTAC-IFBench.
Experimental results reveal substantial deficiencies in multi-turn instruction-following: 
Even the leading LLM, GLM-5.2, fails to
follow about 20\% of the process constraints, while most LLMs only achieve perfect constraint compliance in fewer than 10\% of turns.
In-depth analysis reveals several critical insights:
(1) Instruction-following ability of all LLMs drops rapidly as the interaction grows longer, with near-complete failures in perfectly following all constraints in later turns.
(2) Perfectly following process constraints is significantly more challenging than functional requirements, indicating that final-outcome evaluation can overlook critical failures during the development process.
(3) Code agents perform well on isolated static constraints (e.g., file encoding and placement), but struggle with more complex constraints (e.g., multi-step coordination, project-wide consistency, and precise quantitative control), suggesting their weakness in maintaining global state and disciplined workflows over long interactions.
(4) Code agents show stronger long-horizon robustness for constraints in repository policy files than user instructions. Meanwhile, in user instructions, updating existing constraints is notably harder than adding new ones, indicating difficulty in tracking evolving user intent. 
These findings establish MTAC-IFBench as a valuable resource for advancing reliable instruction-following in multi-turn agentic coding.

%% file: tables/comparison_wo_color.tex
\begin{table*}[!t]
\centering
\resizebox{\linewidth}{!}{
\begin{tabular}{@{}lccccccc@{}}
\toprule
\multirow{2}{*}{Benchmark}  & \multicolumn{3}{c}{Dataset Size}  & Agentic & Instruction & Constraint & \multirow{2}{*}{Evaluation Method} \\
\cmidrule(lr){2-4}
 & \#Inst. & \#Turn & \#Cons. & Scenario & Following & Taxonomy & \\
\midrule
SWE-Bench~\cite{jimenez2024swe}    & 2,294  & 1.00  & 0 & \ding{51} & \ding{55}  & - & Code \\
SWE-Together~\cite{swetogether}    & 109  & -  & 0 & \ding{51} & \ding{55}  & - & Agent-as-a-Judge \\
SWE-\textsc{Interact}~\cite{sweinteract}    & 75  & -  & 0 & \ding{51} & \ding{55}  & - & Code \\
Terminal-Bench~\cite{merrill2026terminal}  & 89  & 1.00 & 0 & \ding{51} & \ding{55}  & - & Code \\
\textsc{ComplexBench}~\cite{wen2024benchmarking} & 1,150 & 1.00 & 4.19 & \ding{55} & \ding{51}  & 4-19 & Code \& LLM-as-a-Judge \\
CodeIF~\cite{yan2025codeif} & 1,200  & 1.00 & 12.90 & \ding{55} & \ding{51} & 8 & Code \\
CodeIF-Bench~\cite{wang2025codeif} & 124 & 7.77  & 7.77 & \ding{55} & \ding{51}  & 9 & Code \\
\textsc{AgentIF}~\cite{qi2025agentif} & 707 & 1.00   & 11.90 & \ding{51} & \ding{51}  & 3 & Code \& LLM-as-a-Judge \\
OctoBench~\cite{ding2026octobench} & 217 & 1.11   & 32.71 & \ding{51} & \ding{51}  & - & LLM-as-a-Judge \\
\midrule
\textbf{MTAC-IFBench (ours)}   & \textbf{100} & \textbf{7.04}  & \textbf{91.33} & \ding{51} & \ding{51} & \textbf{6-18} & \textbf{Code \& Agent-as-a-Judge} \\ 
\bottomrule
\end{tabular}
}
\caption{Comparisons between MTAC-IFBench and other benchmarks. 
The dataset size contains the number of instances (\#Inst.), the average turns (\#Turn) per instance, and the average number of constraints (\#Cons.) per instance. SWE-Together and SWE-\textsc{Interact} utilize user simulators to determine the number of interaction turns dynamically.}
\label{tab:comparisons}
\end{table*}

%% file: sections/related_work.tex
\paragraph{Evaluation of Code Generation.}
As LLMs are increasingly deployed in complex code generation scenarios, the evaluation paradigm for this task is continuously transforming \cite{jiang2026survey}. 
While early benchmarks mainly focused on function \cite{chen2021evaluating, austin2021program, hendrycks2021measuring} or class-level \cite{du2023classeval}, recent research has increasingly shifted toward repository-level and environment-backed tasks \cite{deng2025swe, wu2025frontalk, zhou2026featurebench, li2024deveval}. 
For instance, SWE-Bench \cite{jimenez2024swe} evaluates LLMs on resolving real-world software issues by generating codebase-level patches, while Terminal-Bench \cite{merrill2026terminal} assesses LLMs on long-horizon command-line tasks in terminal environments.
Furthermore, SWE-Together \cite{swetogether} and SWE-\textsc{Interact} \cite{sweinteract} attempt to evaluate repository-level code generation in multi-turn interactive settings.
These benchmarks require LLMs to act as autonomous code agents that leverage project-wide context, iteratively utilize tools, and interact with environments, providing realistic testbeds for code generation in production.
Nevertheless, these benchmarks remain outcome-oriented, evaluating performance by functional correctness while overlooking instruction and constraint compliance during code generation, a crucial aspect of real-world software development \cite{ding2026octobench}.

\paragraph{Evaluation of Instruction-Following.}
The ability to follow instructions is a critical factor determining the practicality of LLMs \cite{liu2023trustworthy, lou2024large},
prompting many benchmarks to evaluate it from various aspects.
Earlier works mainly focus on instruction-following in general chat \cite{zhou2023instruction, wen2024benchmarking, zhang2025cfbench} and single-file code generation \cite{yan2025codeif, wang2025codeif}.
Subsequent works, such as AgentIF \cite{qi2025agentif}, AgentIF-Oneday \cite{chen2026agentif}, and CCTU \cite{ye2026cctu}, expand evaluations to more complex agentic and tool-use contexts. 
However, confined to non-coding or single-turn scenarios, they struggle to assess instruction-following in agentic coding. 
Although OctoBench \cite{ding2026octobench} pioneers instruction-following evaluation for repository-grounded coding agents by synthesizing multi-source constraints, it lacks a systematic constraint taxonomy and is largely confined to single-turn instructions, hindering comprehensive evaluation.
Consequently, MTAC-IFBench is proposed to robustly evaluate code agents' instruction-following ability in multi-turn scenarios.

%% file: sections/methods.tex
\begin{figure*}[!t]
\scriptsize
    \centering
    \includegraphics[width=1\textwidth]{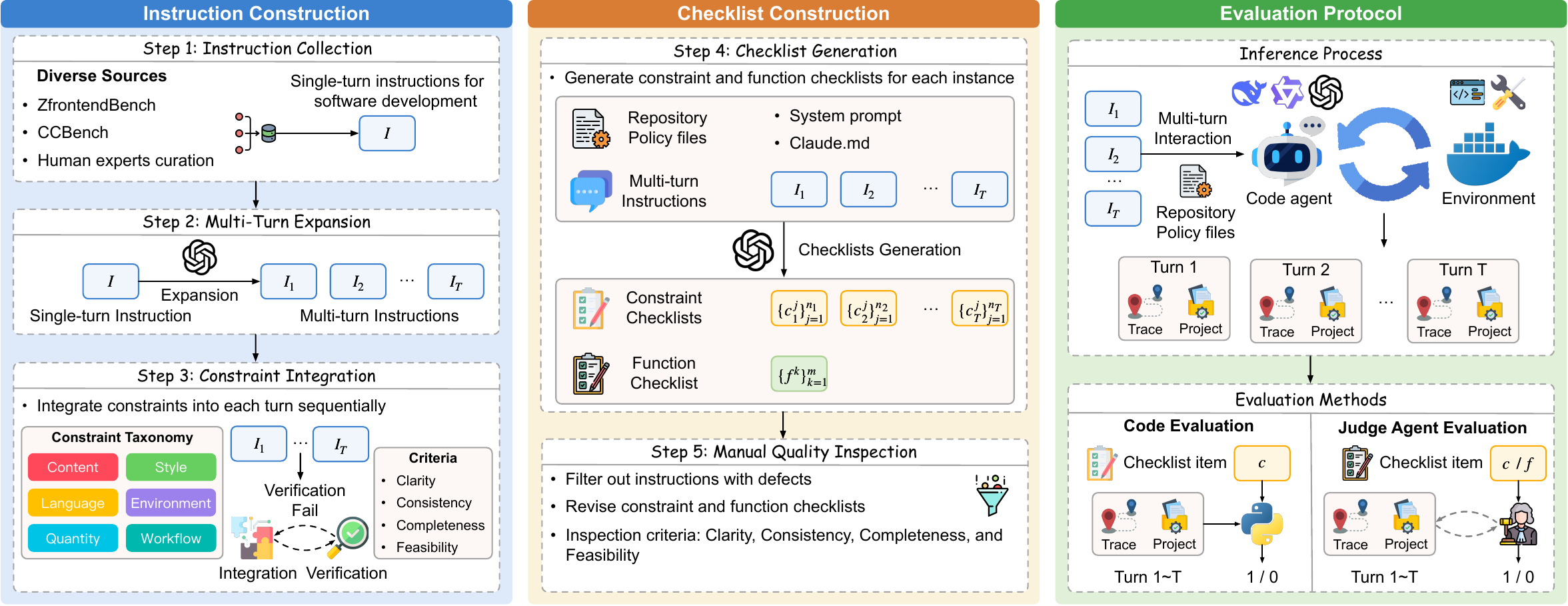}
    \caption{The data construction process and evaluation workflow of MTAC-IFBench. \textit{Left}: Construction of multi-turn software development instructions with diverse constraints. \textit{Center}: Function and constraint checklist curation, as well as manual data quality inspection. \textit{Right}: Evaluation workflow of code agent.}
    \label{fig:overview}
\end{figure*}

In this section, we provide a detailed introduction to MTAC-IFBench, including the following four components: the constraint taxonomy (§ \ref{constraint_category}), the dataset construction process (§ \ref{dataset_construction}), the dataset statistics (§ \ref{dataset_statistics}), and the evaluation
protocol (§ \ref{evaluation_protocol}).
Figure \ref{fig:overview} illustrates the dataset construction and evaluation workflow of MTAC-IFBench.

\subsection{Constraint Taxonomy}
\label{constraint_category}

To comprehensively evaluate LLMs' ability to follow instructions in multi-turn agentic coding, we systematically investigate instructions derived from real-world application scenarios\footnote{These scenarios originate from an LLM-based agentic coding service platform serving over one million users daily.} and existing instruction-following benchmarks for both coding and agentic settings \cite{yan2025codeif, wang2025codeif, yang2025ifevalcode, qi2025agentif, ding2026octobench} to construct a hierarchical constraint taxonomy.
This taxonomy comprises 6 primary and 18 secondary categories, underpinning the construction of diverse and challenging test cases.
Further details and examples are in Appendix \ref{app:details_of_constraint_taxonomy}.

\begin{itemize}[leftmargin=*]
\item \textbf{Content Constraints} regulate the information and formats within the generated artifacts. 
These constraints ensure that agents' outputs align with designated communication norms and data schemas, which include:  
1) \textbf{Keyword} requires or forbids the presence of lexical elements within the generated code or responses.
2) \textbf{Persona} specifies the role, perspective, or tone the agent must adopt to ensure the interaction remains contextually appropriate.
3) \textbf{Format} mandates that outputs adhere to structural templates or markup schemas (e.g., JSON, XML, or boilerplate injection).

\item \textbf{Language Constraints} govern the linguistic and encoding specifications for the generated artifacts, ensuring consistency across multilingual or cross-platform codebases.
This category covers: 1) \textbf{Response Language} for responses or user-facing documentation, 2) \textbf{Comment Language} for code comments, and 3) \textbf{File Encoding} for specified character encodings (e.g., UTF-8) of created files.

\item \textbf{Quantity Constraints} impose numerical boundaries on the generated artifacts, challenging the fine-grained control ability of code agents. 
Specifically, these constraints enforce a designated 1) \textbf{Range} or 2) \textbf{Exact Value} for quantitative attributes such as character counts, lines of code, and comment lengths. 
Furthermore, they restrict the 3) \textbf{Complexity} of the project's architectural and logical density, capping metrics such as directory hierarchy levels, loop nesting depth, and cyclomatic complexity.

\item \textbf{Style Constraints} govern the visual presentation and engineering practices of the generated code, which are crucial for collaborative repositories and engineering standards, including:
1) \textbf{Layout} regulates the spatial organization of code elements, such as indentation, spacing, and comment placement, to maintain consistent readability.
2) \textbf{Naming} specifies naming conventions of variables, functions, classes, and files.
3) \textbf{Paradigm} constrains the programming style and engineering strategy, such as requiring a specific programming pattern and restricting over-engineering.

\item \textbf{Environment Constraints} regulate file-level interactions with repositories and runtime environments, ensuring safe and traceable project modifications in agentic coding.
This category includes:
1) \textbf{File Path} specifies rules for referencing or placing files.
2) \textbf{File Operation} restricts how agents manipulate files, such as creating, modifying, deleting, or backing up.
3) \textbf{Logging} requires records of development procedures, such as modification plans, change logs, or review notes.

\item \textbf{Workflow Constraints} govern the dynamic execution and lifecycle of the agentic coding process, ensuring efficient and verifiable operations during autonomous development.
This category includes:
1) \textbf{Tool Usage} dictates the selection, argument passing, execution rules, and safety wrappers of tools.
2) \textbf{Orchestration} enforces the sequencing and coordination of multi-step actions, such as task planning and parallel execution.
3) \textbf{Testing} regulates code verification practices, specifying required testing frameworks, coverage metrics, and the management of verification records.
\end{itemize}

\subsection{Dataset Construction}
\label{dataset_construction}

As shown in Figure \ref{fig:overview}, MTAC-IFBench is constructed through a semi-automated pipeline comprising four components: instruction collection and multi-turn expansion, taxonomy-guided constraint integration, evaluation methods design, and manual quality inspection.

\paragraph{Instruction Collection and Multi-Turn Expansion.}
To obtain diverse and representative instructions for benchmark construction, our initial dataset integrates high-quality software development instructions derived from both human expert curation and existing benchmarks (i.e., ZfrontendBench \cite{zeng2026glm} and CCBench\footnote{\url{https://huggingface.co/datasets/zai-org/CC-Bench-trajectories}}), 
which cover diverse software development domains, spanning frontend development, tool creation, data analysis, testing, and algorithm implementation. 
Given that the original instructions are single-turn, we employ Gemini-3.1-Pro \cite{team2023gemini} to elevate the task complexity and expand the initial instructions into progressive, multi-turn instruction sequences spanning 5 to 10 turns.
Specifically, via an incremental expansion prompt, the model is guided to generate instructions that first request a basic project skeleton, and then progressively introduce new features or modify existing requirements across subsequent turns.
By performing two independent expansions per instruction, we obtain a total of 374 multi-turn instruction sequences.
Detailed prompts are in Appendix \ref{app:prompt_templates}.

\paragraph{Taxonomy-Guided Constraint Integration.}
To integrate constraints into multi-turn instruction sequences, we employ a turn-wise, iterative integration and validation pipeline to ensure constraints' reasonableness and diversity.
Specifically, we first integrate constraints into the repository policy files (e.g., Claude.md), 
and then into the user instructions at each turn.
During each step, we sample several secondary constraint categories from our taxonomy and instruct Seed-2.0-Pro \cite{seed2026seed} to integrate some constraints into fluent and natural instructions, based on corresponding examples for each category.
The sampling algorithm prioritizes underrepresented categories in previous steps for even distribution.
Furthermore, to simulate the dynamic evolution of user requirements, the integration process is allowed to introduce new constraints or modify existing ones from previous turns.
Upon completing the constraint integration at each turn, we instruct Seed-2.0-Pro to validate the clarity, consistency, completeness, and feasibility of the generated instructions for each turn, triggering reintegration for any failures.
Instruction sequences that exceed the maximum retry threshold at any turn are discarded. 
More details are in Appendix \ref{app:details_of_constraint_integration}.

\paragraph{Checklist Generation and Evaluation Methods Design.}
To accurately verify compliance with each process constraint and the correctness of each functional requirement throughout the development lifecycle, Seed-2.0-Pro is instructed to generate a constraint checklist at each turn and a function checklist upon the conclusion of the multi-turn interaction, which comprehensively captures all immediate turn-level constraints and ultimate functional requirements, respectively.
Subsequently, each checklist item is manually assigned one of the evaluation methods based on its characteristics: 
1) \textbf{Code evaluation} utilizes scripts for verification, which is used for objective and deterministic constraints, such as Keyword and Exact Value. 
2) \textbf{Judge agent evaluation} leverages an auxiliary code agent for verification and determines the final judgment through iterative interaction with the code repository or its build artifacts,
which is used for subjective constraints or those necessitating complex verification procedures, such as Paradigm and Orchestration.
Given the complexity of functional requirement verification, we utilize a judge agent to verify all function checklist items.

\paragraph{Manual Quality Inspection.}
To ensure data quality, we recruit 17 college students to manually inspect the synthesized multi-turn instructions for clarity, consistency, completeness, and feasibility. 
Instructions with any defects are strictly filtered out.
For the retained instructions, we further review the corresponding function and constraint checklists to correct any errors, thereby ensuring strict one-to-one alignment with the requirements and constraints of each instruction.
More details of human inspection are in Appendix \ref{app:human_annotation}.

\subsection{Evaluation Protocol}
\label{evaluation_protocol}
As shown in Figure \ref{fig:overview}, 
upon sequentially feeding the multi-turn instructions to the code agent to generate the interaction traces and code project per turn,
we employ the annotated evaluation method to verify each checklist item.
Claude-Opus-4.6 \cite{anthropic2026opus} is used to generate verification scripts for code evaluation, and the judge agent is built on the Claude Code framework with Claude-Sonnet-4.6 \cite{anthropic2026sonnet} for function checklists and DeepSeek-V4-Pro\footnote{\url{https://huggingface.co/deepseek-ai/DeepSeek-V4-Pro}} for constraint checklists.
The judge agent is permitted to use Playwright\footnote{\url{https://github.com/microsoft/playwright}} to simulate browser interactions and capture screenshots for functional verification. 
More details are in Appendix \ref{app:evaluation_protocol}.

\subsection{Dataset Statistics}
\label{dataset_statistics}
Table \ref{tab:comparisons} presents a comparison between MTAC-IFBench and other related benchmarks. 
Specifically, MTAC-IFBench contains 100 instances and 9,133 constraints.
On average, each instance comprises 7.04 interaction turns, 16.07 function checklist items, and 91.33 constraint checklist items (averaging 12.97 constraints per turn). 
This constraint density poses a substantial challenge to the instruction-following abilities of code agents in multi-turn software development scenarios.
Furthermore, MTAC-IFBench covers a diverse spectrum of constraint categories, as shown in Figure \ref{fig:constraint_distribution}.
We also present the task distribution of MTAC-IFBench in Appendix \ref{app:task_distribution}.

\begin{figure}[!t]
  \centering
  \includegraphics[width=1.0\linewidth]{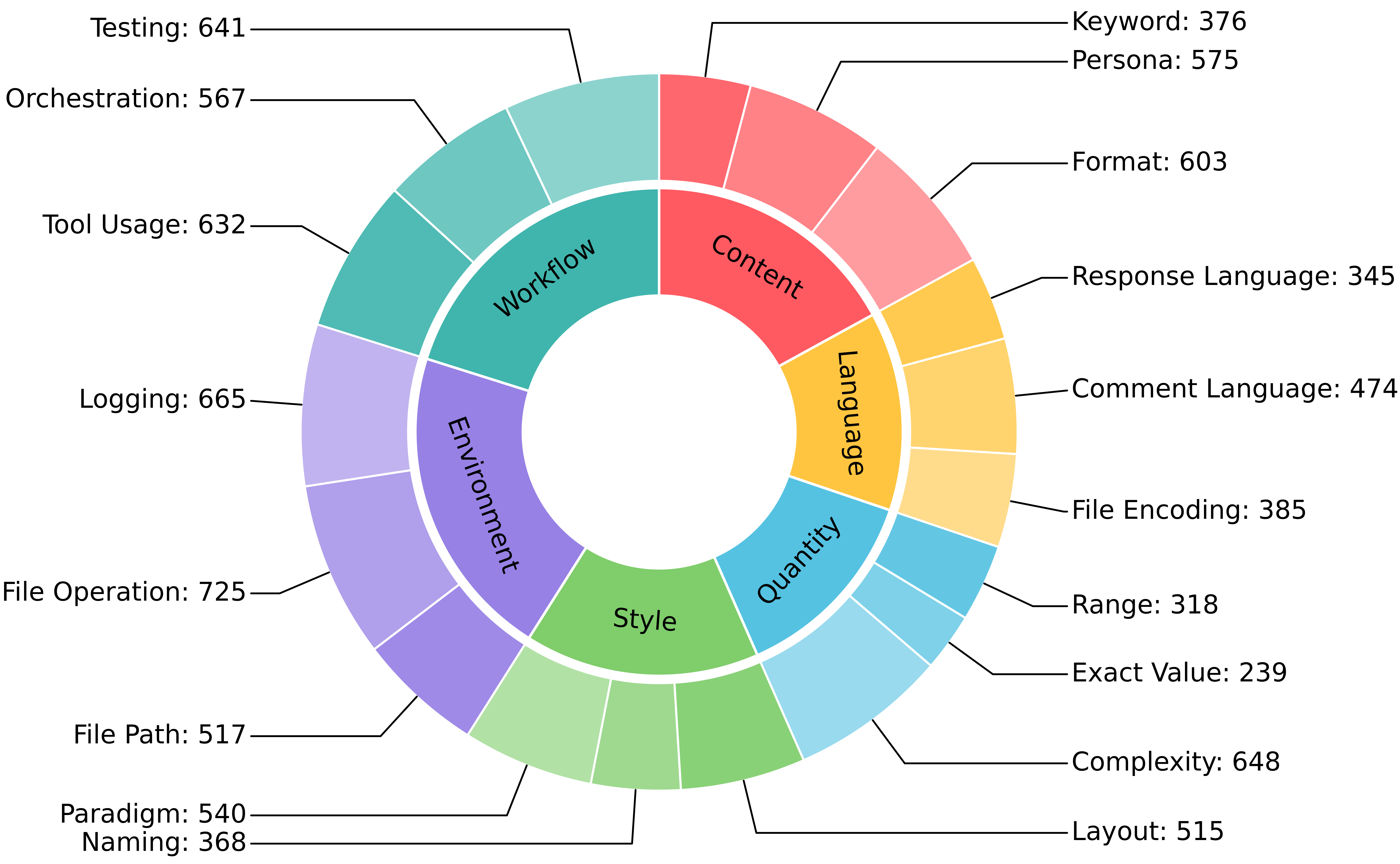}
  \caption{The distribution of constraints in MTAC-IFBench.}
  \vspace{-4mm}
  \label{fig:constraint_distribution}
\end{figure}

%% file: sections/experiments.tex
\input{tables/instruction_following_results}
\subsection{Evaluation Metrics}
We evaluate both turn-level instruction-following and final functional correctness. 
For a multi-turn instruction $\{I_1, \dots, I_T\}$, let $\mathcal{C}_t$ denote the constraint checklist at turn $t$ and $z_{t,j} \in \{0,1\}$ denote whether the $j$-th constraint is satisfied. 
Upon completion of the final turn, let $b \in \{0,1\}$ denote the executability of the generated project, $\mathcal{F}$ denote the function checklist, and $y_k \in \{0,1\}$ denote whether the $k$-th function is satisfied.
Following previous work \cite{qin2025sysbench, qi2025agentif, ding2026octobench}, we define the following evaluation metrics. For instruction-following, \textbf{CSR$_t$} (Constraint Success Rate) measures the proportion of satisfied constraints at turn $t$, while \textbf{C-ISR$_t$} (Constraint-Instruction Success Rate) measures strict completion where all constraints must be satisfied. 
For final functional correctness, \textbf{BSR} (Build Success Rate) measures whether the final generated project is executable, \textbf{FSR} (Function Success Rate) measures the proportion of satisfied functions, and \textbf{F-ISR} (Function-Instruction Success Rate) measures strict completion where all functional requirements must be satisfied. 
For each instance, these metrics are computed as follows:
\begingroup
\setlength{\abovedisplayskip}{6pt}
\setlength{\belowdisplayskip}{6pt}
\[
\begin{gathered}
\begin{alignedat}{4}
&\mathrm{CSR}_t &&= \frac{1}{|\mathcal{C}_t|}\sum_{j=1}^{|\mathcal{C}_t|} z_{t,j} 
&\qquad &\mathrm{C\text{-}ISR}_t &&= \prod_{j=1}^{|\mathcal{C}_t|} z_{t,j} \\[6pt]
&\mathrm{FSR}   &&= \frac{1}{|\mathcal{F}|}\sum_{k=1}^{|\mathcal{F}|} y_k 
&\qquad &\mathrm{F\text{-}ISR}   &&= \prod_{k=1}^{|\mathcal{F}|} y_k
\end{alignedat}
\\[6pt]
\mathrm{BSR} = b
\end{gathered}
\]
\endgroup
For a dataset with $N$ instances, the final reported metrics are obtained by averaging these instance-level scores.

\subsection{Evaluation Settings}
We evaluate 11 advanced LLMs on MTAC-IFBench, including 5 powerful proprietary LLMs: Claude-Opus-4.6 \cite{anthropic2026opus}, Gemini-3.1-Pro \cite{gemini}, Qwen3.6-Plus \cite{qwen3.6}, Seed-2.0-Pro \cite{seed2026seed}, Claude-Haiku-4.5 \cite{anthropic2025haiku}, as well as 6 strong open-source LLMs: GLM-5.2 \cite{zeng2026glm}, GLM-5.1, DeepSeek-V4-Pro, DeepSeek-V4-Flash\footnote{\url{https://huggingface.co/deepseek-ai/DeepSeek-V4-Flash}}, Kimi-K2.6 \cite{kimi}, and Qwen3.5-27B\footnote{\url{https://huggingface.co/Qwen/Qwen3.5-27B}}.
For all models, we employ Claude Code (v2.1.14) \cite{claude} as the standard agent harness and retain the default decoding parameters of their respective APIs. 
We also analyze the performance of each model under the OpenCode (v1.1.21) \cite{opencode} framework (§ \ref{sec:analysis}).

\input{tables/functional_correctness_results}

\subsection{Main Results}
%\paragraph{Main Results.}
Tables \ref{tab:instruction_following_results} and \ref{tab:functional_correctness_results} detail LLMs' performance on instruction-following and functional correctness, respectively. 
\textbf{Firstly}, MTAC-IFBench poses a rigorous challenge to current LLM-based code agents. 
Even the leading LLM, GLM-5.2, fails to follow about 20\% of the constraints. 
Although LLMs achieve moderate CSR ranging from 60.6\% to 80.4\%, their C-ISR remains remarkably low, mostly falling below 10\%. 
This exposes the immense difficulty of perfectly complying with procedural instructions in agentic coding scenarios.
\textbf{Secondly}, instruction-following performance degrades substantially as the interaction horizon grows, evidenced by the continuous decline of both CSR and C-ISR across all models. 
For most models, the C-ISR even plummets to zero after six turns. 
This highlights the challenges of instruction-following in multi-turn interactions and aligns with our motivation.
Nevertheless, stronger models exhibit greater multi-turn robustness, showing substantially smaller performance degradation as the number of turns increases. 
\textbf{Finally}, despite a strong positive correlation between the performance of instruction-following and functional correctness, the C-ISR of all models remains significantly lower than their F-ISR, indicating that perfectly satisfying procedural constraints is more difficult than functional requirements and exposing the limitations of outcome-oriented evaluations.
% \textbf{Finally}, open-source models are competitive with proprietary models in MTAC-IFBench. 
% The leading open-source LLM, GLM-5.1, closely approaches the lead proprietary model, Claude-Opus-4.6, in both instruction-following and functional correctness, demonstrating its potential as a viable alternative to closed-source LLMs.

\begin{figure}[!t]
  \centering
  \includegraphics[width=1.0\linewidth]{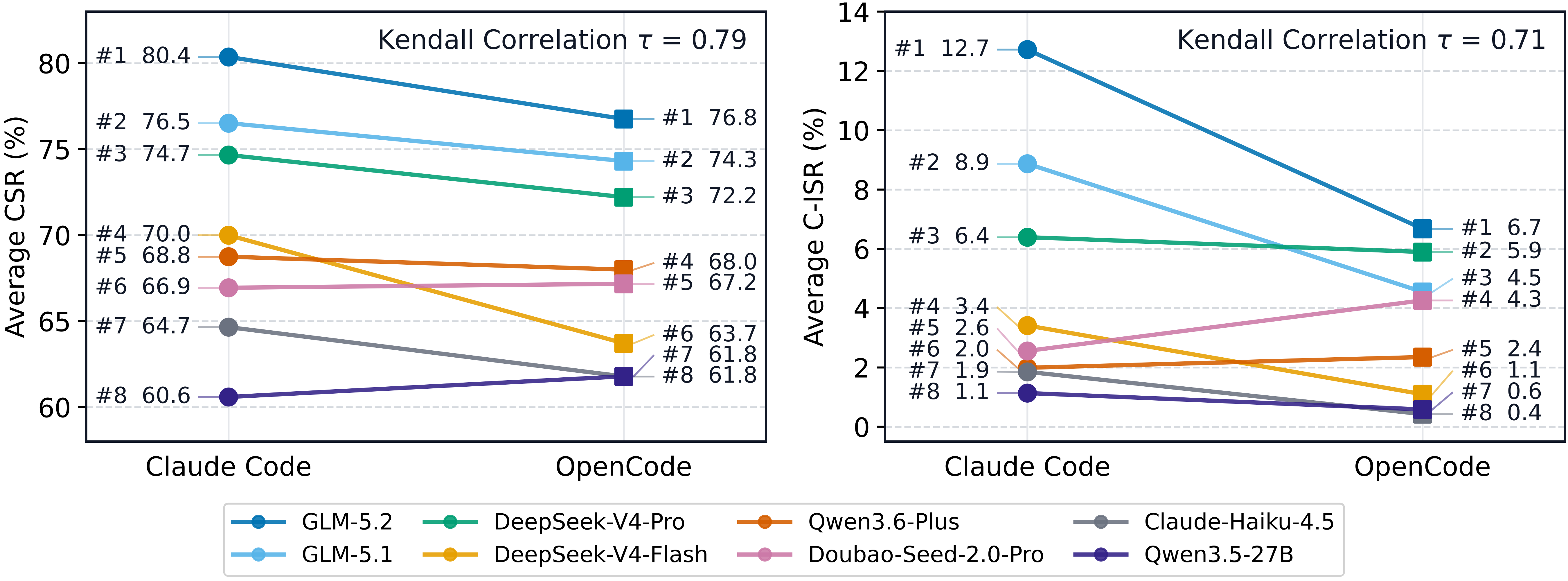}
  \caption{Average CSR and C-ISR of LLMs under the Claude Code and OpenCode framework. }
  \label{fig:harness_comparison}
\end{figure}

\begin{figure}[!t]
  \centering
  \includegraphics[width=1.0\linewidth]{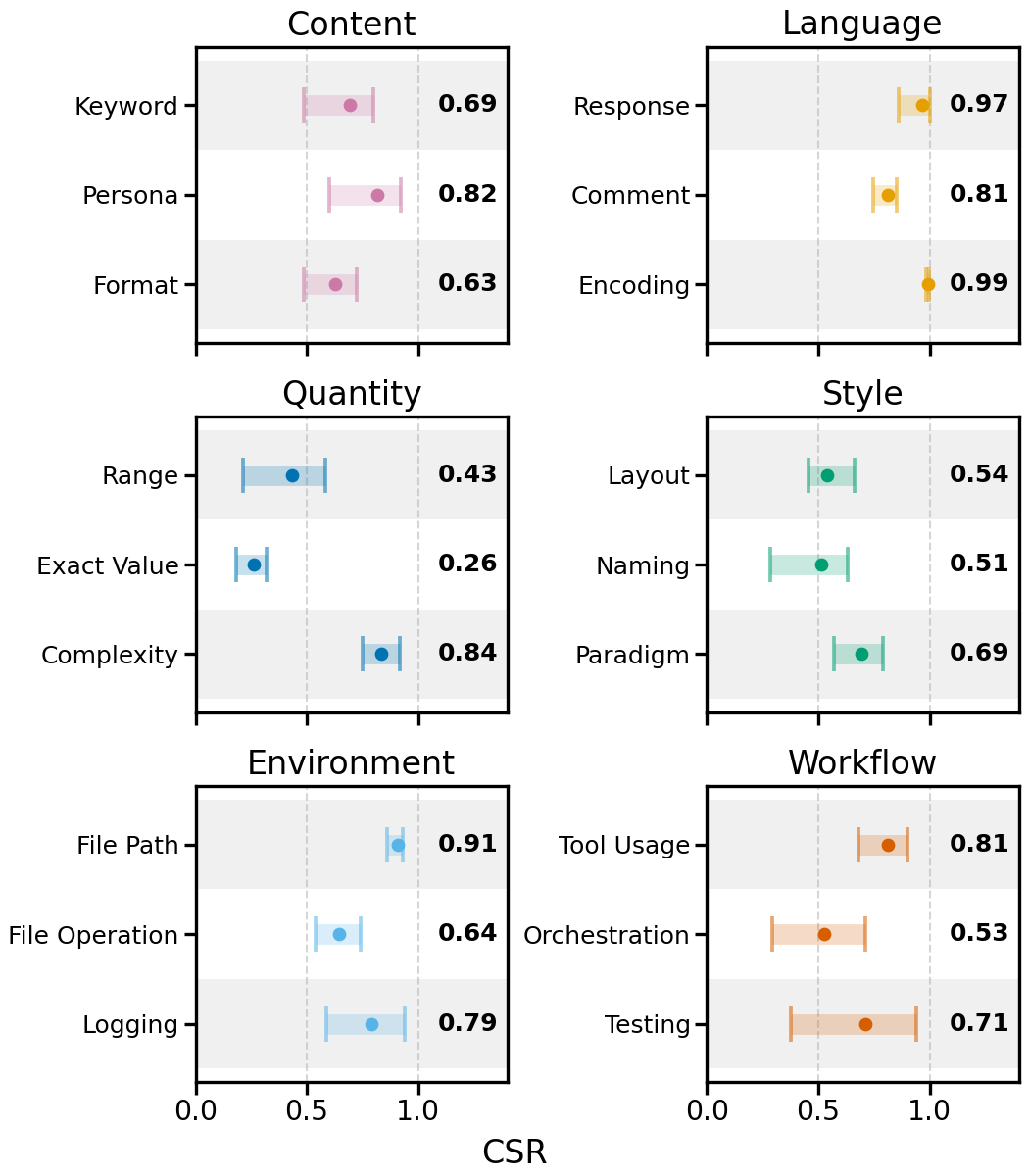}
  \caption{The mean and range of CSR across all LLMs on different constraint categories.}
  \label{fig:constraint_from_categories}
\end{figure}

\subsection{Analysis}
\label{sec:analysis}
\paragraph{Performance on Different Agent Harnesses.}
To examine the impact of agent harnesses on instruction-following performance, we compare eight LLMs under two widely used frameworks, Claude Code and OpenCode, in Figure \ref{fig:harness_comparison}.
\textbf{Firstly}, models generally achieve stronger instruction-following performance under Claude Code, which tends to produce higher CSR and C-ISR scores.
For example, GLM-5.2's CSR drops from 80.4\% under Claude Code to 76.8\% under OpenCode, while its C-ISR drops more sharply from 12.7\% to 6.7\%.
This demonstrates the substantial impact of the harness on code-agent performance.
\textbf{Secondly}, the broad model hierarchy remains largely stable across different harnesses, particularly among the top-performing models, as reflected by the strong Kendall correlations.
However, fine-grained rankings among middle- and lower-performing models still exhibit greater harness-specific variation, highlighting the importance of standardizing and explicitly reporting the agent harness when evaluating code agents.
More detailed experimental results and analyses are in Appendix \ref{app:experiment_detail}, where we observe a similar performance decay under OpenCode as the interaction horizon grows, consistent with our main results.

\paragraph{Performance on Different Constraint Categories.}
To identify which constraints are most difficult for code agents, we report the mean and range of CSR across all LLMs on different constraint categories in Figure \ref{fig:constraint_from_categories}.
\textbf{Firstly}, multi-turn orchestration and maintaining global project consistency pose severe challenges. 
Unlike simple one-off code generation, agentic coding requires models to manage complex workflow actions (e.g., Orchestration) and enforce repository-wide style standards across different files and turns (e.g., Paradigm and Naming). 
The low CSR in these two categories demonstrates that agents easily lose context and struggle to maintain architectural cohesion during extended interactions.
\textbf{Secondly}, code agents exhibit a critical vulnerability in precise quantitative control. 
The remarkably low success in Quantity constraints indicates a severe lack of the strict state-tracking mechanisms required for fine-grained code adjustments.
\textbf{Finally}, there is a striking disparity between agents' handling of static rules versus dynamic workflows. 
While LLMs achieve near-perfect compliance on isolated constraints such as Encoding or File Path, their performance sharply declines on constraints requiring continuous planning and global coordination. 
This demonstrates that merely generating compliant static code is vastly insufficient for acting as an independent agent capable of navigating dynamic and state-complex real-world projects.

\begin{figure}[!t]
  \centering
  \includegraphics[width=1.0\linewidth]{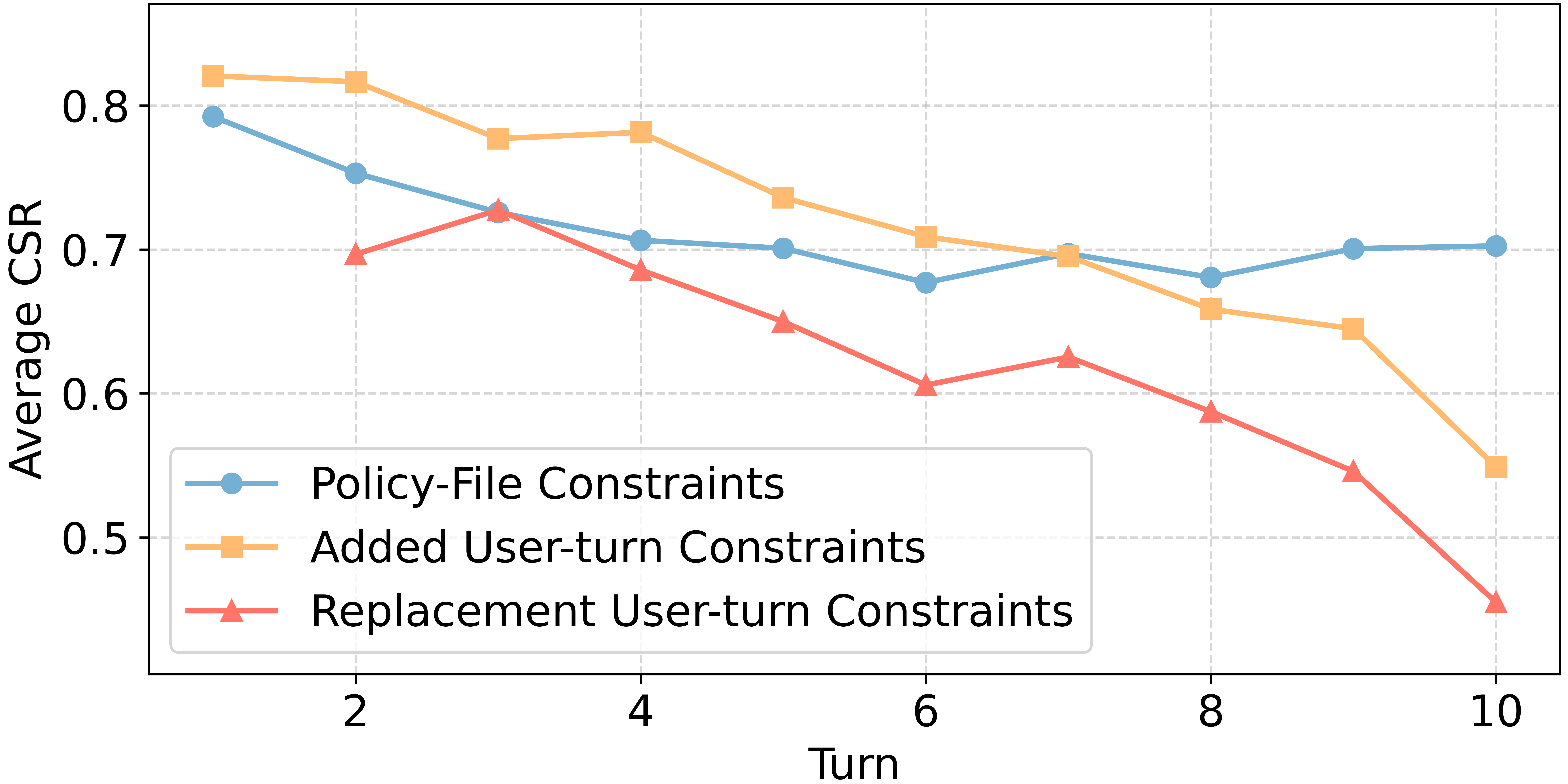}
  \caption{Average CSR of constraints from different sources across all LLMs. Since there are no constraint replacements in Turn 1, this result starts from Turn 2.}
  \label{fig:constraint_from_sources}
\end{figure}

\paragraph{Performance on Different Constraint Sources.} 
To dissect the performance of code agents across constraint sources, we compare the average CSR for three constraint sources: repository policy files, along with newly added constraints and updates replacing existing constraints within user instructions. 
The results are presented in Figure \ref{fig:constraint_from_sources}.
\textbf{Firstly}, code agents show stronger long-horizon robustness for constraints in repository policy files, exhibiting significantly less performance decay over successive turns compared to those in user instructions.
This finding suggests that keeping important constraints in the repository policy file remains a more reliable practice for code agents.
\textbf{Secondly}, updating existing constraints is notably harder than adding new ones. 
This likely stems from interference by superseded constraints, highlighting the inherent difficulty of dynamically revising requirements in multi-turn interactions.

\paragraph{Evaluation Reliability.}
To validate the reliability of our evaluation method, we randomly sample 200 function checklist items and 400 constraint checklist items (164 evaluated by code and 236 evaluated by judge agents) for manual assessment. 
Two annotators independently verify compliance for each item, while a third inspector conducts cross-validation. 
Any discrepancies are resolved through discussion to reach a consensus. 
The agreement between the automatic evaluation and the manual assessment is shown in Table \ref{tab:human_consistency}.
For constraint checklist items, both evaluation methods achieve an impressive agreement of over 92\%. 
Although verifying function checklist items is significantly more challenging, our method still reached an 85.5\% agreement, surpassing the 85\% agreement between the two initial annotators. 
These results confirm the reliability of our evaluation methods.

\input{tables/human_consistency}

%% file: tables/instruction_following_results.tex
\begin{table*}[!t]
\centering
\resizebox{\textwidth}{!} {
\begin{tabular}{l|ccccc|c|ccccc|c}
\toprule
\textbf{Metrics} & \multicolumn{6}{c|}{\textbf{CSR}} & \multicolumn{6}{c}{\textbf{C-ISR}} \\
\midrule
\multirow{2}{*}{\textbf{Model}} & \multicolumn{5}{c|}{\textbf{Turns}} & \multirow{2}{*}{\textbf{Avg.}} & \multicolumn{5}{c|}{\textbf{Turns}} & \multirow{2}{*}{\textbf{Avg.}} \\
\cmidrule{2-6} \cmidrule{8-12}
 & 1 $\sim$ 2 & 3 $\sim$ 4 & 5 $\sim$ 6 & 7 $\sim$ 8 & 9 $\sim$ 10 & & 1 $\sim$ 2 & 3 $\sim$ 4 & 5 $\sim$ 6 & 7 $\sim$ 8 & 9 $\sim$ 10 & \\
\midrule
\multicolumn{13}{c}{\textit{Proprietary language models}} \\
\midrule
Claude-Opus-4.6 & 82.4 & 79.1 & 76.8 & 76.4 & \textbf{76.2} & 78.6 & 19.0 & 6.5 & 3.5 & 3.3 & 2.3 & 8.7 \\
Gemini-3.1-Pro  & 81.6 & 74.9 & 67.7 & 61.7 & 58.2 & 71.0 & \textbf{28.1} & 9.1 & 2.4 & 0.0 & 0.0 & 11.1 \\
Qwen3.6-Plus    & 73.3 & 68.3 & 67.2 & 67.6 & 65.0 & 68.8 & 6.0 & 1.0 & 0.0 & 0.0 & 0.0 & 2.0 \\
Seed-2.0-Pro    & 75.6 & 68.8 & 63.8 & 61.9 & 57.6 & 66.9 & 8.0 & 1.0 & 0.0 & 0.0 & 0.0 & 2.6 \\
Claude-Haiku-4.5 & 72.4 & 67.8 & 62.5 & 59.8 & 49.7 & 64.7 & 4.5 & 2.0 & 0.0 & 0.0 & 0.0 & 1.9 \\
\midrule
\multicolumn{13}{c}{\textit{Open-source language models}} \\
\midrule
GLM-5.2 & \textbf{85.8} & \textbf{80.8} & \textbf{77.9} & \textbf{78.0} & 76.1 & \textbf{80.4} & 27.6 & \textbf{9.7} & \textbf{4.8} & \textbf{5.8} & \textbf{2.6} & \textbf{12.7} \\
GLM-5.1  & 81.3 & 77.9 & 74.7 & 71.5 & 72.2 & 76.5 & 19.7 & 8.2 & 2.6 & 0.0 & 0.0 & 8.9 \\
DeepSeek-V4-Pro & 79.6 & 76.9 & 72.7 & 70.8 & 67.7 & 74.7 & 14.5 & 7.0 & 1.2 & 0.0 & 0.0 & 6.4 \\
Kimi-K2.6       & 80.7 & 74.7 & 70.5 & 70.1 & 62.5 & 73.2 & 18.6 & 7.2 & 1.8 & 0.0 & 0.0 & 7.8 \\
DeepSeek-V4-Flash & 76.4 & 71.7 & 68.7 & 65.6 & 59.6 & 70.0 & 8.5 & 2.0 & 1.2 & 1.1 & 0.0 & 3.4 \\
Qwen3.5-27B     & 68.3 & 62.6 & 57.7 & 56.5 & 50.5 & 60.6 & 4.0 & 0.0 & 0.0 & 0.0 & 0.0 & 1.1 \\
\bottomrule
\end{tabular}
}
\caption{The instruction-following performance of LLMs under the Claude Code framework. CSR and C-ISR are bucketed by turn intervals to compute average scores. 
All metrics are reported as percentages (\%). The best result is \textbf{bold}.}
\vspace{-4mm}
\label{tab:instruction_following_results}
\end{table*}

%% file: tables/functional_correctness_results.tex
\begin{table} [!t]
\centering
\small
\begin{tabular}{l|c|c|c}
\toprule
\textbf{Model} & \textbf{BSR} & \textbf{FSR} & \textbf{F-ISR}  \\
\midrule
% \multicolumn{4}{c}{\textit{Proprietary language models}} \\
% \midrule
Claude-Opus-4.6 & \textbf{98.0} & \textbf{90.2} & \textbf{37.0} \\
Gemini-3.1-Pro  & 90.0 & 66.5 & 14.0 \\
Qwen3.6-Plus    & 96.0 & 66.5 & 9.0 \\
% \midrule
% \multicolumn{4}{c}{\textit{Open-source language models}} \\
% \midrule
GLM-5.1         & \textbf{98.0} & 83.0 & 29.0 \\
DeepSeek-V4-Pro & 97.0 & 79.2 & 26.0 \\
Kimi-K2.6       & 97.0 & 75.2 & 29.0 \\
\bottomrule
\end{tabular}

\caption{The functional correctness of LLMs. 
All metrics are reported as percentages (\%). 
The best result is \textbf{bold}.}
\label{tab:functional_correctness_results}
\end{table}

%% file: tables/human_consistency.tex
\begin{table} [!t]
\centering
\resizebox{\linewidth}{!} {
\begin{tabular}{l|c|c|c|c}
\toprule
\textbf{Category} & \textbf{Evaluation Method} & \textbf{Agr.} & \textbf{P-F1} & \textbf{N-F1}  \\
\midrule
\multirow{2}{*}{Constraint Checklist}  & Code & 0.921 & 0.939 & 0.887 \\
    & Judge Agent & 0.924 & 0.949 & 0.845 \\
\midrule
Function Checklist     & Judge Agent  & 0.855 & 0.895 & 0.764 \\
\bottomrule
\end{tabular}
}
\caption{Agreement, Positive F1 scores (P-F1), and Negative F1 scores (N-F1) between our evaluation methods and manual assessment for checklist item verification.}
\label{tab:human_consistency}
\end{table}

%% file: sections/conclusion.tex
Our work introduces MTAC-IFBench, a comprehensive benchmark for instruction-following in multi-turn agentic coding scenarios. 
We first propose a hierarchical constraint taxonomy of agentic coding encompassing 6 primary and 18 secondary categories. 
A semi-automated pipeline is then employed to construct multi-turn software development instructions with diverse constraints. 
For evaluation, we construct checklists for turn-wise constraint compliance and final functional correctness, integrating verification scripts and judge agents to ensure evaluation reliability.
Experimental results uncover the significant deficiencies of code agents in multi-turn instruction-following, as the performance degrades rapidly as the interaction horizon grows. 
In summary, MTAC-IFBench can serve as a practical tool to advance future research in instruction-following for code agents.

%% file: sections/appendix.tex
\section{Limitations}
\input{sections/limitations}

\section{Ethical Considerations}
\input{sections/ethical_considerations}

\section{Details of Constraint Taxonomy}
\label{app:details_of_constraint_taxonomy}
We provide detailed descriptions and examples for each constraint category in Table \ref{tab:constraint_taxonomy}.

\section{List of Prompt Templates}
\label{app:prompt_templates}
This section lists all the prompt templates applied
throughout this work, including the prompt for multi-turn instruction expansion in Table \ref{tab:multi_turn_expansion_prompt}, the prompt for constraint integration into the repository policy files in Table \ref{tab:add_constraints_first_turn_prompt}, the prompt for constraint integration into the user instruction at each turn in Table \ref{tab:add_constraints_multi_turn_prompt}, the prompt for constraint quality validation at each turn in Table \ref{tab:instruction_quality_evaluation_prompt}, the prompt for function checklist generation in Table \ref{tab:checklist_generation_prompt}, 
the prompt for verification code generation in Table \ref{tab:validation_code_generation_prompt}, 
the prompt for constraint checklist verification in Table \ref{tab:instruction_following_verification_prompt}, and the prompt for function checklist verification in Table \ref{tab:function_checklist_verification_http} and \ref{tab:function_checklist_verification_non_http}.

\section{Details of Constraint Integration}
\label{app:details_of_constraint_integration}
Before constraint integration, we first construct a diverse constraint example pool for each secondary constraint category. 
Specifically, we first manually collect 190 representative constraint examples across all constraint categories by investigating real-world application scenarios and existing instruction-following benchmarks in both coding and agentic settings, and then employ Seed-2.0-Pro to rephrase these examples, generating 10 variations for each example, ensuring an adequate volume for each secondary category.

For repository policy files, the constraint integration process begins with an empty file.
We sample 4-6 secondary constraint categories and retrieve up to 10 examples per category.
Using these examples, Seed-2.0-Pro is instructed to select or rewrite up to 8 constraints and integrate them into the repository policy file.

For each turn of user instruction, we sample 2-4 secondary constraint categories, again retrieving up to 10 examples per category.
Additionally, we supply constraint examples related to constraints from previous turns, allowing Seed-2.0-Pro to introduce new constraints or modify existing ones.
The total number of new or modified constraints is capped at two per turn.
To avoid ambiguity caused by conflicts between repository policy files and user instructions, constraints in repository policy files are kept immutable throughout the interaction; only constraints introduced in previous turns of user instruction are allowed to be modified.

When sampling secondary categories, we track an exposure count for each category, defined as the cumulative number of turns its constraints remain active in the constructed dataset so far.
Categories are ranked by these counts, and sampling weights are assigned based on the square of their reverse ranks.
This approach assigns higher probabilities to less exposed categories, promoting a balanced distribution of constraint categories. 
By default, every integrated constraint applies to all subsequent turns unless explicitly modified.

After each integration step, we use Seed-2.0-Pro to validate the clarity, consistency, completeness, and feasibility of the generated instruction.
If validation fails, the integration step is retried using the same sampled examples.
If no valid instruction is obtained after 5 attempts, the entire instruction sequence is discarded.

\section{Details of Manual Quality Inspection}
\label{app:human_annotation}
\paragraph{Inspection Guidelines}
For instruction quality inspection, we provide annotators with the repository policy files and multi-turn instructions, asking them to inspect the clarity, consistency, completeness, and feasibility of these instructions. 
Instructions with any defects are strictly filtered out.
The detailed guideline is shown in Table \ref{tab:instruction_quality_inspection_guideline}.
For checklist revision, we provide annotators with the retained repository policy files, the multi-turn instructions, the corresponding functional constraint checklists, as well as the constraint checklists at each turn, asking them to check the correctness of these checklists and correct any errors.
The detailed guideline is shown in Table \ref{tab:checklist_revision_guideline}.

\paragraph{Profile of Inspection Persons}
In the manual quality inspection of MTAC-IFBench, we recruit a diverse group of 17 highly qualified annotators who are either holding or pursuing a bachelor's degree from top universities.
This cohort includes 11 Master's and 6 Bachelor's graduates in Computer Science, Statistics, or Electrical Engineering.   
All of them are required to pass the mandatory proficiency examination and annotation tutorial.
Additionally, 5 professional data engineers and the authors are tasked with conducting spot checks on the annotations.

\paragraph{Quality Assurance and Validation}
We establish a rigorous quality control workflow to ensure the high quality of our annotations. 
The annotation results for each instance are independently reviewed by an inspector. 
Any discrepancies are resolved through discussion between the inspector and the annotator to reach a consensus on the final annotation. 
Furthermore, we regularly monitor the annotation quality and provide feedback to the annotators, while the annotation protocols undergo iterative refinement through collective discussion to enhance clarity and precision.

\paragraph{Data Curation Cost}
We spend approximately 2,400\$ on the curation
of MTAC-IFBench during the manual quality inspection step.

\input{tables/instruction_following_results_opencode}
\input{tables/task_distribution}

\section{Details of Evaluation Protocol}
\label{app:evaluation_protocol}
\paragraph{Code Evaluation}
For code evaluation, the software development project generated by the code agent and the verification script are placed in the same execution environment. 
The project path and the code agent's final-turn response are passed as inputs to the verification script.
Based on these inputs, the verification script provides a binary judgment regarding whether the constraint is followed.
The verification code generation prompt is in Table \ref{tab:validation_code_generation_prompt}.

\paragraph{Judge Agent Evaluation}
For the evaluation of constraint checklist items, the judge agent is executed within the same environment as the software development project generated by the code agent. 
The operation trace of the evaluated code agent, its final-turn response, and the project directory are provided to the judge agent. Based on these inputs, the judge agent is expected to output a binary judgment for each item on the constraint checklist.
During this step, we employ the prompting strategy of \textsc{IF-Critic} \cite{wen2025if}, which requires the judge model to evaluate all constraints within a given checklist during a single inference pass. 
This approach has been proven to be cost-efficient and enables the judge agent to achieve superior evaluation performance.
The evaluation prompt is in Table \ref{tab:instruction_following_verification_prompt}.
For the evaluation of the function checklist items, the judge agent is also executed within the same environment as the software development project generated by the code agent. 
The project directory is provided to the judge agent, and the judge agent is permitted to use Playwright\footnote{\url{https://github.com/microsoft/playwright}} to simulate browser interactions and capture screenshots for functional verification. 
For frontend development projects, we additionally employ npm\footnote{\url{https://www.npmjs.com/}} for automated deployment and provide the accessible HTTP URL to the judge agent. 
The judge agent is built on the Claude Code (v2.1.14) \cite{claudecode} framework for both scenarios.
The evaluation prompts are detailed in Table \ref{tab:function_checklist_verification_http} and \ref{tab:function_checklist_verification_non_http}.

\section{Task Distribution of MTAC-IFBench}
\label{app:task_distribution}
We adopt the taxonomy from CC-Bench\footnote{\url{https://huggingface.co/datasets/zai-org/CC-Bench-trajectories}} to categorize the task types of instructions in MTAC-IFBench. 
As shown in Table \ref{tab:task_distribution}, MTAC-IFBench encompasses diverse software development tasks, spanning frontend development, application development, and data analysis, among others.

\section{Details of Experiments}
\label{app:experiment_detail}
\begin{figure}[!t]
  \centering
  \includegraphics[width=1.0\linewidth]{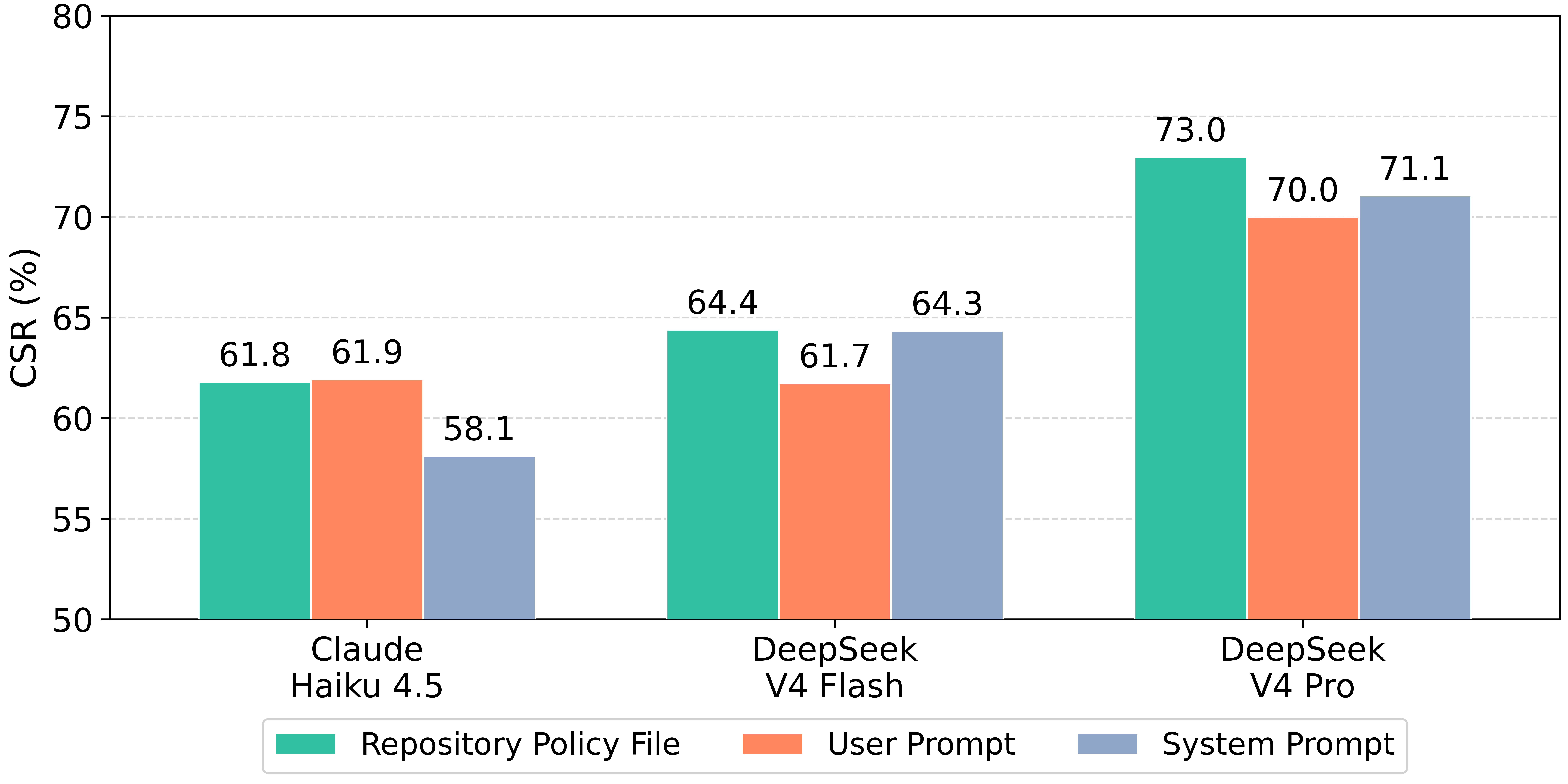}
  \caption{Average CSR of constraints from repository policy file when they are placed in different positions. }
  \label{fig:system_prompt_from_sources}
\end{figure}

\paragraph{Detailed Results under OpenCode Framework.} Table \ref{tab:instruction_following_results_opencode} presents the performance of different LLMs under the OpenCode framework across interaction turns.
Consistent with our observations under Claude Code, model performance steadily declines as the interaction horizon grows.
Although models generally perform worse under OpenCode than under Claude Code, they exhibit stronger multi-turn robustness in OpenCode, with less
performance decay over successive turns.
This result indicates the advantage of 
context management mechanism in OpenCode.

\paragraph{Performance on Different Constraint Placements.}
To investigate how constraint placement affects instruction-following performance,
we place the constraints from the repository policy file in two different locations: the system prompt and the first-turn user prompt. 
Due to cost considerations, we randomly sample 30 instances from MTAC-IFBench for this experiment.
The average CSR of these constraints across all turns is shown in Figure \ref{fig:system_prompt_from_sources}.
We observe that constraints placed in the repository policy file typically achieve the best compliance, while placement in the system prompt and user prompt shows similar performance. Therefore, keeping important constraints in the repository policy file remains a best practice for code agents.

\input{tables/constraint_category}
\input{tables/multi_turn_expansion_prompt}
\input{tables/constraint_intergration_first_turn_prompt}
\input{tables/constraint_integration_multi_turn_prompt}
\input{tables/quality_validation_prompt}
\input{tables/checklist_generation_prompt}
\input{tables/instruction_quality_inspection_guideline}
\input{tables/checklist_revision_guideline}
\input{tables/verification_code_generation_prompt}
\input{tables/instruction_following_verification_prompt}
\input{tables/function_verification_prompt_http}
\input{tables/function_verification_prompt_other}

%% file: sections/limitations.tex
The limitations of our work are summarized as follows:

\paragraph{Potential Evaluation Bias.} 
Similar to other model-based evaluation methods, MTAC-IFBench may also suffer from potential evaluation biases such as self-enhancement or verbosity bias \cite{zheng2023judging}.
These biases could harm the evaluation correctness of certain models.
Although we have verified a high agreement rate between our evaluation methods and human expert evaluations, incorporating inference-time strategies (e.g., multiple judge agents debate) may further attenuate bias and enhance evaluation performance. 
We reserve further investigation of evaluation bias mitigation as important future work.

\paragraph{Cost of Judge Agent.}
MTAC-IFBench adopts judge agents to verify constraint and functional requirements that cannot be reliably checked by static scripts.
Although this design enables more faithful evaluation of complex requirements, it also introduces additional cost because judge agents may need multiple rounds of tool calls and environment interaction.
We consider this a necessary trade-off for evaluating multi-turn agentic coding, and leave further cost reduction through more dedicated designs, such as utilizing more lightweight agent harnesses or base models, and parallelizing the verification of multiple requirements, as important future work.

\paragraph{Coverage of Development Scenarios.}
Although MTAC-IFBench has covered diverse software development tasks, real-world agentic coding scenarios may further involve broader task types, such as security vulnerability remediation and database schema migration. 
Further expanding the coverage of instruction scenarios can improve the robustness of the evaluation and is regarded as important future work.

%% file: sections/ethical_considerations.tex
In this work, we recruit 17 college students for benchmark data quality inspection, as detailed in Appendix \ref{app:human_annotation}.
Throughout the data annotation process, we adhere to the following key principles: (1) All annotators are fully informed about the purpose of the study, the specific tasks involved, and the intended use of their annotated data. 
(2) All annotators receive fair compensation for their time and contributions based on the market price.
(3) All annotators are granted full autonomy to withdraw from the project at any time without penalty.
(4) Any harmful instructions are filtered before annotation to avoid ethical issues.

Regarding the investigation of instructions derived from real-world application scenarios, we only use the portion of data that is granted for research purposes by users and conduct a strict deidentification and desensitization process to protect user privacy.

%% file: tables/instruction_following_results_opencode.tex
\begin{table*}[!t]
\centering
\resizebox{\textwidth}{!} {
\begin{tabular}{l|ccccc|c|ccccc|c}
\toprule
\textbf{Metrics} & \multicolumn{6}{c|}{\textbf{CSR}} & \multicolumn{6}{c}{\textbf{C-ISR}} \\
\midrule
\multirow{2}{*}{\textbf{Model}} & \multicolumn{5}{c|}{\textbf{Turns}} & \multirow{2}{*}{\textbf{Avg.}} & \multicolumn{5}{c|}{\textbf{Turns}} & \multirow{2}{*}{\textbf{Avg.}} \\
\cmidrule{2-6} \cmidrule{8-12}
 & 1 $\sim$ 2 & 3 $\sim$ 4 & 5 $\sim$ 6 & 7 $\sim$ 8 & 9 $\sim$ 10 & & 1 $\sim$ 2 & 3 $\sim$ 4 & 5 $\sim$ 6 & 7 $\sim$ 8 & 9 $\sim$ 10 & \\
\midrule
\multicolumn{13}{c}{\textit{Proprietary language models}} \\
\midrule
Qwen3.6-Plus    & 71.8 & 69.8 & 66.4 & 64.4 & 59.6 & 68.0 & 7.0 & 0.0 & 1.0 & 0.0 & 0.0 & 2.4 \\
Seed-2.0-Pro    & 76.0 & 70.2 & 64.1 & 60.5 & 56.3 & 67.2 & 14.0 & 1.0 & 0.0 & \textbf{1.0} & 0.0 & 4.3 \\
Claude-Haiku-4.5 & 67.3 & 64.4 & 60.2 & 57.8 & 51.3 & 61.8 & 2.0 & 0.0 & 0.0 & 0.0 & 0.0 & 1.0 \\
\midrule
\multicolumn{13}{c}{\textit{Open-source language models}} \\
\midrule
GLM-5.2 & \textbf{80.6} & \textbf{77.3} & \textbf{75.3} & \textbf{74.1} & \textbf{74.2} & \textbf{76.8} & \textbf{15.0} & 4.5 & \textbf{4.7} & 0.0 & 0.0 & \textbf{6.7} \\
GLM-5.1  & 77.9 & 75.1 & 72.4 & 72.0 & 70.9 & 74.3 & 11.5 & 3.8 & 0.0 & 0.0 & 0.0 & 4.5 \\
DeepSeek-V4-Pro & 75.6 & 74.5 & 71.1 & 67.8 & 65.0 & 72.2 & 11.3 & \textbf{7.3} & 1.6 & 0.0 & 0.0 & 5.9 \\
DeepSeek-V4-Flash & 68.5 & 64.9 & 62.2 & 61.4 & 54.4 & 63.7 & 3.2 & 1.0 & 0.0 & 0.0 & 0.0 & 1.1 \\
Qwen3.5-27B     & 66.4 & 62.4 & 58.3 & 59.7 & 59.6 & 61.8 & 2.0 & 0.0 & 0.0 & 0.0 & 0.0 & 1.0 \\
\bottomrule
\end{tabular}
}
\caption{The instruction-following performance of LLMs under the OpenCode framework. CSR and C-ISR are bucketed by turn intervals to compute average scores. 
All metrics are reported as percentages (\%). The best result is \textbf{bold}.}
\label{tab:instruction_following_results_opencode}

\end{table*}

%% file: tables/task_distribution.tex
\begin{table}[t]
\centering
\begin{tabular}{lr}
\toprule
\textbf{Category} & \textbf{\# Instances} \\
\midrule
Frontend Development & 24 \\
Application Development & 20 \\
UI/UX Optimization & 19 \\
Build \& Deployment & 10 \\
Data Analysis & 22 \\
Machine Learning & 5 \\
\midrule
Total & 100 \\
\bottomrule
\end{tabular}
\caption{Task distribution of MTAC-IFBench.}
\label{tab:task_distribution}
\end{table}

%% file: tables/constraint_category.tex
\begin{table*}[t]
\small
\centering
\resizebox{\linewidth}{!} {
\renewcommand{\arraystretch}{1.12}
\setlength{\tabcolsep}{1.3mm}{
    \begin{tabular}{@{}l|l|>{\raggedright\arraybackslash}m{0.40\textwidth}|>{\raggedright\arraybackslash}m{0.30\textwidth}@{}}
    \toprule
    \textbf{Primary Category} & \textbf{Secondary Category} & \textbf{Description} & \textbf{Example} \\
    \specialrule{\lightrulewidth}{0pt}{0pt}
    
    \multirow{6.5}{*}{Content}
    & Keyword & Constraints that require or forbid the presence of lexical elements within the generated code or responses. & Every code file you generate must contain \texttt{"TODO"}. \\
    \cline{2-4}
    & Persona & Constraints that specify the role, perspective, or conversational manner the agent should consistently adopt. & Every response must end with the modal particle \texttt{"meow\textasciitilde"}. \\
    \cline{2-4}
    & Format & Constraints that require generated artifacts to follow a designated structural template or markup schema. &
    Every code file you generate must add the encoding declaration on the first line: \texttt{-*- coding: utf-8 -*-}. \\
    
    \specialrule{\lightrulewidth}{0pt}{0pt}
    
    \multirow{4.5}{*}{Language}
    & Response Language & Constraints that specify the language used for responses or user-facing documentation. & You must always generate user responses in English. \\
    \cline{2-4}
    & Comment Language & Constraints that specify the language used for code comments. & All comments in every code file you generate must be in Chinese. \\
    \cline{2-4}
    & File Encoding & Constraints that regulate the character encodings (e.g., UTF-8) of created files. & Every code file you create must be encoded in UTF-8. \\
    \specialrule{\lightrulewidth}{0pt}{0pt}
    
    \multirow{6}{*}{Quantity}
    & Range & Constraints that enforce designated ranges for quantitative attributes of generated artifacts. & Each file you generate must contain between 200 and 500 lines. \\
    \cline{2-4}
    & Exact Value & Constraints that enforce quantitative attributes to match exact numerical values or discrete conditions. & Every code file you generate must contain exactly 4 comment lines. \\
    \cline{2-4}
    & Complexity & Constraints that cap the architectural or logical density of generated projects and code. & The directory hierarchy of the code project you generate must not exceed 2 levels. \\
    \specialrule{\lightrulewidth}{0pt}{0pt}
    
    \multirow{5}{*}{Style}
    & Layout & Constraints that regulate the spatial organization of code elements. & The code files you generate must use 4 spaces for indentation. \\
    \cline{2-4}
    & Naming & Constraints that specify naming conventions for variables, functions, classes, or files. & Every function name in the code files you generate must follow snake case. \\
    \cline{2-4}
    & Paradigm & Constraints that specify the programming style or engineering strategy used in implementation. & The code you generate must avoid adding error handling, fallbacks, or validation for impossible scenarios. \\
    \specialrule{\lightrulewidth}{0pt}{0pt}
    
    \multirow{6.5}{*}{Environment}
    & File Path & Constraints that specify rules for referencing or placing files in the repository. & In the code files you generate, all file references must use absolute file paths instead of relative paths. \\
    \cline{2-4}
    & File Operation & Constraints that restrict how agents create, read, modify, delete, or back up files. & Before modifying an existing file each time, you must create a backup of the original file in the same directory. \\
    \cline{2-4}
    & Logging & Constraints that require traceable records of development procedures, plans, or changes. & Before every code modification, you must write the modification plan to \texttt{task\_plan.md}. \\
    \specialrule{\lightrulewidth}{0pt}{0pt}
    
    \multirow{8}{*}{Workflow}
    & Tool Usage & Constraints that dictate tool selection, argument passing, execution rules, or safety wrappers. & You must use dedicated tools, such as Read, Write, and Edit, as much as possible for file operations instead of bash commands. \\
    \cline{2-4}
    & Orchestration & Constraints that enforce the sequencing and coordination of multi-step agent actions. & You must execute multiple independent tool calls in parallel as much as possible to improve efficiency. \\
    \cline{2-4}
    & Testing & Constraints that regulate code verification practices, coverage requirements, or verification records. & The code you generate must be checked with ESLint, and all errors must be fixed so that ESLint finally returns no errors. \\
    
    \bottomrule
    \end{tabular}
}
}
\caption{Constraint taxonomy of MTAC-IFBench.}
\label{tab:constraint_taxonomy}
\end{table*}

%% file: tables/multi_turn_expansion_prompt.tex
\begin{table*}[!t]
\small
\centering
\setlength{\tabcolsep}{1.6mm}{
\begin{tabular}{p{\dimexpr \linewidth-2\tabcolsep\relax}}
\toprule
You are an expert specializing in constructing multi-turn software development instructions. 
Given a brief description of a project, you need to decompose and refine it into a progressive multi-turn instruction sequence.
\newline
\newline
\#\# Core Requirements
\newline
1. \textbf{First turn (turn\_id=0)}: Outline the basic skeleton or draft of the entire project.
\newline
\quad - Set up the project framework, create the basic page, or build the basic structure.
\newline
\quad - You may include one or a few very simple interactive elements, such as a button or a placeholder page, but do not implement the core business functionality.
\newline
\quad - Leave sufficient room for extension in later turns.
\newline
\newline
2. \textbf{Subsequent turns (turn\_id=1, 2, ...)}: In each turn, add \textbf{one or a few} sub-feature requirements based on the previous turns. The requirements should be as complex and challenging for the model as reasonably possible.
\newline
\quad - Each turn should focus on only one or a few functionally similar sub-features; do not include too many tasks in a single turn.
\newline
\quad - Except for the first turn, the functional complexity or level of each turn should be roughly comparable.
\newline
\quad - Instructions in later turns may refer to, modify, or override content from previous turns.
\newline
\newline
3. \textbf{Instruction Style Requirements}:
\newline
\quad - Use a natural multi-turn dialogue style, such as ``Good, next...'', ``Now let's improve...'', or ``The next step is...''.
\newline
\quad - The description in each turn must be clear, accurate, and written from the user's perspective; it is not necessary to restate the original specifications from earlier turns, item by item.
\newline
\quad - Do not write descriptions that are overly abstract or impossible to implement. Instead, write concrete and implementable feature requirements.
\newline
\newline
4. \textbf{Number of turns}: Split the project reasonably according to its complexity, aturn \textcolor{red}{\{turn\_num\}} turns.
\newline
\newline
\#\# Output Format
\newline
Please strictly output the complete result in the following JSON format. You must ensure that the JSON structure is complete and properly closed, do not omit any element, and do not include any other text:
\newline
\newline
{[}
\newline
\hspace*{2em}\{
\newline
\hspace*{4em}``turn\_id'': 0,
\newline
\hspace*{4em}``instruction'': ``The specific instruction for the first turn...''
\newline
\hspace*{2em}\},
\newline
\hspace*{2em}\{
\newline
\hspace*{4em}``turn\_id'': 1,
\newline
\hspace*{4em}``instruction'': ``The specific instruction for the second turn...''
\newline
\hspace*{2em}\},
\newline
\hspace*{2em}...
\newline
{]}
\newline
\newline
\#\# Project Description
\newline
\textcolor{red}{\{description\}}
\newline
\newline
Please output the decomposed multi-turn instruction sequence in JSON format.
\\
\bottomrule
\end{tabular}
}
\caption{The prompt template for multi-turn instruction expansion.}
\label{tab:multi_turn_expansion_prompt}
\end{table*}

%% file: tables/constraint_intergration_first_turn_prompt.tex
\begin{table*}[!t]
\scriptsize
\linespread{0.80}\selectfont
\centering
\setlength{\tabcolsep}{1.6mm}{
\begin{tabular}{p{\dimexpr \linewidth-2\tabcolsep\relax}}
\toprule
You are an expert specializing in adding new constraints to existing instructions. 
I will provide you with an instruction for a code agent. 
Your task is to add appropriate constraints to the instruction, thereby producing a new instruction that integrates both the original instruction and the added constraints. All available constraints and examples are provided below:
\newline
\newline
\textbf{[The Start of Available Constraints]}
\newline
\textcolor{red}{\{constraint\_examples\}}
\newline
\textbf{[The End of Available Constraints]}
\newline
\newline
You need to follow the rules below:
\newline
1. You can freely select multiple constraints to add to the instruction, but the template of each newly constructed constraint must be taken exactly from one of the examples above. When synthesizing constraints, you must satisfy the following requirements:
\newline
\hspace*{1em} a. Each newly added constraint must be obtained by replacing placeholders in a constraint template. Only minor wording adjustments that do not affect the semantics are allowed when necessary to make the constraint integrate smoothly with the instruction.
\newline
\hspace*{1em} b. Each newly added constraint must be sufficiently challenging for the code agent. You must ensure that, without this constraint, the code agent has a reasonably high probability of violating it, and after adding this constraint, the code agent needs to make additional adjustments to its response.
\newline
\hspace*{1em} c. Same constraint templates may be added multiple times. For the keyword constraint, multiple keywords may be specified at the same time when semantically reasonable, so as to further increase the difficulty of the constraint.
\newline
\hspace*{1em} d. Maintain diversity when adding constraints, and do not limit yourself to only a few constraint types. You may place relatively more emphasis on constraints related to tool usage, file operations, and the overall workflow.
\newline
\hspace*{1em} e. The number of constraints added in this turn should not exceed 8.
\newline
2. The newly added constraints should be natural and reasonable. There must be no contradiction or conflict among the constraints.
\newline
3. When describing these constraints in the instruction, you should clearly state that they must be followed throughout all subsequent turns, so that the code agent responding to the instruction will not mistakenly think that the constraints only need to be followed in the current turn.
\newline
4. The instruction I provide may be empty. In this case, you must not fabricate any task requirements for the instruction. 
Directly append the constraints that need to be followed in this turn at the end of the instruction.
\newline
5. You must ensure that the newly constructed instruction fully satisfies the following requirements:
\newline
\hspace*{1em} a. The intent is very clear.
\newline
\hspace*{1em} b. The semantics are very clear, the language is fluent and easy to understand, and there is no ambiguity, difficult-to-understand wording, or grammatical issue.
\newline
\hspace*{1em} c. There is no incorrect content or false information.
\newline
\hspace*{1em} d. There are no unreasonable constraints. Note that unconventional constraints or constraints that violate common coding conventions are allowed, as long as they are not unreasonable or impossible to complete.
\newline
\hspace*{1em} e. There is no erroneous logic or confused logic.
\newline
\hspace*{1em} f. The internal logic is consistent. There are no contradictory constraints, and there is no conflict between the instruction content and the constraints; all of them can be satisfied at the same time.
\newline
6. The code agent has permission to use tools and perform file operations, so operations such as writing files are feasible.
\newline
\newline
Below is the instruction.
\newline
\textbf{[The Start of the Instruction]}
\newline
\textcolor{red}{\{instruction\}}
\newline
\textbf{[The End of the Instruction]}
\newline
\newline
You need to strictly output the final result in the following format:
\newline
```
\newline
\textbf{[The Start of Analysis]}
\newline
... (You need to analyze and plan how to add appropriate constraints to the instruction.)
\newline
\textbf{[The End of Analysis]}
\newline
\newline
\textbf{[The Start of New Instruction]}
\newline
... (You need to provide the new instruction that integrates the original instruction and the added constraints.)
\newline
\textbf{[The End of New Instruction]}
\newline
\newline
\textbf{[The Start of Existing Constraints]}
\newline
```json
\newline
{[}
\newline
\hspace*{2em}\{
\newline
\hspace*{4em}``constraint\_template'': ... (String; Provide the constraint template on which the first constraint constructed in the new instruction is based. It must be copied exactly without any modification),
\newline
\hspace*{4em}``constraint\_content'': ... (String; Provide the specific content of this constraint in the new instruction. It must be an original text fragment from the new instruction and must be copied exactly without any modification),
\newline
\hspace*{4em}``constraint\_parameters'': ... (Dictionary; Provide the parameter content used by the first constraint constructed in the new instruction. Its format must be exactly the same as the constraint\_parameters field of the example constraint on which it is based)
\newline
\hspace*{2em}\},
\newline
\hspace*{2em}... (And so on, listing the information for each constraint constructed in the new instruction)
\newline
{]}
\newline
```
\newline
\textbf{[The End of Existing Constraints]}
\newline
```
\\
\bottomrule
\end{tabular}
}
\caption{The prompt template for constraint integration into the repository policy files.}
\label{tab:add_constraints_first_turn_prompt}
\end{table*}

%% file: tables/constraint_integration_multi_turn_prompt.tex
\begin{table*}[!t]
\scriptsize
\linespread{0.80}\selectfont
\centering
\setlength{\tabcolsep}{1.6mm}{
\begin{tabular}{p{\dimexpr \linewidth-2\tabcolsep\relax}}
\toprule
You are an expert specializing in adding new constraints to existing instructions. I will provide you with an instruction for a code agent, as well as the constraints that already exist in the instruction. 
The instruction contains a multi-turn dialogue. 
Your task is to add appropriate constraints to the final-turn instruction, thereby producing a more complex new final-turn instruction. All available constraints and examples are provided below:
\newline
\newline
\textbf{[The Start of Available Constraints]}
\newline
\textcolor{red}{\{constraints\}}
\newline
\textbf{[The End of Available Constraints]}
\newline
\newline
You need to follow the rules below:
\newline
1. You can freely select multiple constraints to add to the instruction, but the template of each newly constructed constraint must be taken exactly from one of the examples above. When synthesizing constraints, you must satisfy the following requirements:
\newline
\hspace*{1em} a. Each newly added constraint must be obtained by replacing placeholders in a constraint template. Only minor wording adjustments that do not affect the semantics are allowed when necessary to make the constraint integrate smoothly with the instruction.
\newline
\hspace*{1em} b. Each newly added constraint must be sufficiently challenging for the code agent. You must ensure that, without this constraint, the code agent has a reasonably high probability of violating it, and after adding this constraint, the code agent needs to make additional adjustments to its response.
\newline
\hspace*{1em} c. Same constraint templates may be added multiple times. For the keyword constraint, multiple keywords may be specified at the same time when semantically reasonable, so as to further increase the difficulty of the constraint.
\newline
\hspace*{1em} d. Maintain diversity when adding constraints, and do not limit yourself to only a few constraint types. You may place relatively more emphasis on constraints related to tool usage, file operations, and the overall workflow.
\newline
2. In addition to adding new constraints, in this turn, you can also replace existing constraints from the previous turns with another constraint. You should maintain an appropriate balance between adding new constraints and replacing existing constraints. 
The \textbf{[Existing Constraints]} include all constraints that need to be satisfied from the previous context. 
However, only constraints in the \textbf{[Editable Constraints]} can be replaced. 
Constraints not in the \textbf{[Editable Constraints]} must be fully preserved in the final output under \textbf{[Existing Constraints]} without deletion, modification, or reordering.
\newline
4. The final-turn instruction only needs to describe the constraints that are newly added or replaced in this turn. Existing constraints from the previous turns do not need to be repeated. The total number of newly added and replaced constraints in this turn should not exceed two.
\newline
5. Except for minor wording adjustments made when adding new constraints to ensure fluency, do not make any modifications to the original instruction.
\newline
6. The newly added constraints should be natural and reasonable. There must be no contradiction or conflict among the constraints.
\newline
7. When describing these constraints in the instruction, you should clearly state that they must be followed throughout all subsequent turns, so that the code agent responding to the instruction will not mistakenly think that the constraints only need to be followed in the current turn.
\newline
8. After constructing the new final-turn instruction, you should provide:
\newline
\hspace*{1em} a. \textbf{All existing constraints} that need to be satisfied when answering the final-turn instruction, including constraints that appeared in the previous turns and are not replaced. If you choose to replace a constraint when constructing the new instruction, the replaced old constraint should no longer be treated as an existing constraint.
\newline
\hspace*{1em} b. All constraints newly added or replaced in this turn. The number should not exceed two, and the constraint should use the actual content in this turn's instruction rather than the content from previous turns. Every constraint listed under \textbf{[Constraints Newly Added or Replaced in This Turn]} must also appear in the final output under \textbf{[Existing Constraints]}.
\newline
9. The code agent has permission to use tools and perform file operations, so operations such as writing files are feasible.
\newline
\newline
Below are the original instruction and all constraints that already exist in the original instructions.
\newline
\textbf{[The Start of Instruction History]}
\newline
\textcolor{red}{\{history\}}
\newline
\textbf{[The End of Instruction History]}
\newline
\newline
\textbf{[The Start of Original Instruction]}
\newline
\textcolor{red}{\{final\_turn\_instruction\}}
\newline
\textbf{[The End of Original Instruction]}
\newline
\newline
\textbf{[The Start of Existing Constraints]}
\newline
\textcolor{red}{\{existing\_constraints\}}
\newline
\textbf{[The End of Existing Constraints]}
\newline
\newline
\textbf{[The Start of Editable Constraints]}
\newline
\textcolor{red}{\{editable\_constraints\}}
\newline
\textbf{[The End of Editable Constraints]}
\newline
\newline
You need to strictly output the final result in the following format:
\newline
```
\newline
\textbf{[The Start of Analysis]}
\newline
... (You need to analyze and plan how to add or modify appropriate constraints for the final-turn instruction.)
\newline
\textbf{[The End of Analysis]}
\newline
\newline
\textbf{[The Start of New Instruction]}
\newline
... (You need to provide the new final-turn instruction that integrates the original instruction and the constraints.)
\newline
\textbf{[The End of New Instruction]}
\newline
\newline
\textbf{[The Start of Existing Constraints]}
\newline
\hspace*{1em}```json
\newline
\hspace*{1em}{[}
\newline
\hspace*{2em}\{
\newline
\hspace*{3em}``constraint\_template'': ... (String; Provide the constraint template on which the first constraint is based when answering the final-turn instruction. It must be copied exactly without any modification),
\newline
\hspace*{3em}``constraint\_content'': ... (String; Provide the specific content of this constraint in the new instruction. It must be an original text fragment from the new instruction and must be copied exactly without any modification),
\newline
\hspace*{3em}``constraint\_parameters'': ... (Dictionary; Provide the parameter content used by the first constraint when answering the final-turn instruction. Its format must be exactly the same as the constraint\_parameters field of the example constraint on which it is based)
\newline
\hspace*{2em}\},
\newline
\hspace*{2em}... (And so on, listing the information for each existing constraint that must be satisfied when answering the final-turn instruction)
\newline
\hspace*{1em}{]}
\newline
\hspace*{1em}```
\newline
\textbf{[The End of Existing Constraints]}
\newline
\newline
\textbf{[The Start of Constraints Newly Added or Replaced in This turn]}
\newline
(The output format is the same as \textbf{[Existing Constraints]}, listing the information for each constraint newly added or replaced in this turn)
\newline
\textbf{[The End of Constraints Newly Added or Replaced in This turn]}
\\
\bottomrule
\end{tabular}
}
\caption{The prompt template for constraint integration into the user instruction at each turn.}
\label{tab:add_constraints_multi_turn_prompt}
\end{table*}

%% file: tables/quality_validation_prompt.tex
\begin{table*}[!t]
\small
\centering
\setlength{\tabcolsep}{1.6mm}{
\begin{tabular}{p{\dimexpr \linewidth-2\tabcolsep\relax}}
\toprule
You are an expert specializing in machine learning and natural language processing, and you are skilled at judging the quality of instructions. 
I will provide you with a complex multi-turn instruction, as well as all constraints that must be satisfied when responding to the final-turn instruction. 
Please carefully analyze and scrutinize the logic of the final-turn instruction, and carefully consider whether each constraint in the final-turn instruction can be completed. 
When judging the quality of the instruction, a high-quality instruction must satisfy all of the following conditions. 
If any condition is violated, the instruction should be judged as low quality. 
You need to carefully analyze, think thoroughly, and examine the instruction word by word to determine whether it satisfies each condition below:
\newline
\newline
1. The intent is very clear.
\newline
2. The semantics are very clear, the language is fluent and easy to understand, and there is no ambiguity, difficult-to-understand wording, or grammatical issue.
\newline
3. There is no incorrect content or false information.
\newline
4. There are no unreasonable constraints. 
Note that unconventional constraints that violate common coding conventions are allowed, as long as they are not unreasonable or impossible to complete.
\newline
5. There is no erroneous logic or confused logic.
\newline
6. The internal logic is consistent. There are no contradictory constraints, and there is no conflict between the instruction content and the constraints; all of them can be satisfied at the same time.
\newline
7. The instruction content and materials are complete. All required data or information is provided, and there are no problems such as missing content, missing materials, or missing data.
\newline
\newline
Do not answer the instruction. You only need to judge whether the instruction is high quality or low quality. Strictly follow the format below to output your analysis and conclusion.
\newline
\newline
"""
\newline
Analysis: ... (First provide a complete, comprehensive, detailed, and careful analysis.)
\newline
Final Conclusion: The final conclusion can only be ``[[High Quality]]'' or ``[[Low Quality]]''.
\newline
"""
\newline
\newline
In addition, please pay attention to the following principles:
\newline
1. The artificial intelligence assistant has permission to use tools and perform file operations, so operations such as writing files are feasible.
\newline
2. Constraints may be updated or replaced across different turns, as long as the instruction is sufficiently clear and does not cause ambiguity. In this case, it should not be regarded as a conflict in the instruction content.
\newline
3. The input only provides instructions; the artificial intelligence assistant's responses and written code are not provided. In actual use, the assistant's responses will be obtained interactively, so modifications involving the response or written code content in the instruction are reasonable and should not be regarded as missing materials. It is normal for the first-turn instruction to serve as a system instruction and contain no actual task constraints, but only constraints; you should not judge it as low quality for this reason.
\newline
\newline
The instruction is as follows:
\newline
\textbf{[The Start of the Instruction]}
\newline
\textcolor{red}{\{instruction\}}
\newline
\textbf{[The End of the Instruction]}
\newline
\newline
\textbf{[The Start of Existing Constraints]}
\newline
\textcolor{red}{\{checklist\}}
\newline
\textbf{[The End of Existing Constraints]}
\\
\bottomrule
\end{tabular}
}
\caption{The prompt template for constraint quality validation at each turn.}
\label{tab:instruction_quality_evaluation_prompt}
\end{table*}

%% file: tables/checklist_generation_prompt.tex
\begin{table*}[!t]
\small
\centering
\setlength{\tabcolsep}{1.6mm}{
\begin{tabular}{p{\dimexpr \linewidth-2\tabcolsep\relax}}
\toprule
You are an expert specializing in constructing multi-turn coding task data. Given a project description and a decomposed multi-turn instruction sequence, you need to generate function checklist items for the \textbf{final state} of the project.
\newline
\newline
\#\# Core Requirements
\newline
1. \textbf{Atomicity}: Each checklist item should evaluate only \textbf{one} function. Do not merge multiple functions into a single checklist item.
\newline
2. \textbf{Comprehensive Coverage}: Cover as many implemented features required by all turns of instructions as possible.
\newline
3. \textbf{Clear and Accurate Description}: Clearly specify what operation should be performed, what result is expected, and how to determine whether the behavior is correct or incorrect.
\newline
\newline
\#\# Output Format
\newline
Please strictly output the complete result in the following JSON format. You must ensure that the JSON structure is complete and properly closed, do not omit any element, and do not include any other text:
\newline
\newline
{[}
\newline
\hspace*{2em}\{
\newline
\hspace*{4em}``checklist\_id'': 0,
\newline
\hspace*{4em}``description'': ``The specific checklist description, including operation steps and expected results...'',
\newline
\hspace*{2em}\},
\newline
\hspace*{2em}...
\newline
{]}
\newline
\newline
\#\# Project Description
\newline
\textcolor{red}{\{description\}}
\newline
\newline
\#\# Multi-turn Instruction
\newline
\textcolor{red}{\{multi\_turn\_instruction\}}
\newline
\newline
Please output the checklist items in JSON format.
\\
\bottomrule
\end{tabular}
}
\caption{The prompt template for generating the function checklist upon the conclusion of the multi-turn interaction.}
\label{tab:checklist_generation_prompt}
\end{table*}

%% file: tables/instruction_quality_inspection_guideline.tex
\begin{table*}[t]
\small
\centering
\setlength{\tabcolsep}{1.6mm}{
\begin{tabular}{p{\dimexpr \linewidth-2\tabcolsep\relax}}
\toprule
Below, a multi-turn instruction will be provided. 
The instruction may contain a system prompt and multiple turns of user instructions.
Your task is to determine whether the instruction is high-quality. 
If the instruction contains any of the following defects, it should be filtered out.
\newline
\newline
\textbf{Instruction Quality Definition:} \newline
An instruction consists of the system prompt and all user instructions across turns. A high-quality instruction should be clear, internally consistent, sufficiently complete, feasible to implement, and free from safety, privacy, or confidentiality issues.
\newline
\newline
\textbf{Task Details:}
\newline
\textbf{Step 1: Quality Judgment} \newline
Evaluate the quality of the instruction. Select "Low-quality" if any of the following conditions are satisfied:
\newline
1. Ambiguity: The instruction can be interpreted in multiple substantially different ways, or it is too unclear to infer the expected task or functionality.
\newline
2. Contradiction: Requirements within the same turn or across different turns conflict with each other. Later-turn modifications or replacements of previous requirements should not be treated as contradictions.
\newline
3. Infeasibility: The instruction contains requirements that are unrealistic, impossible to implement, or based on serious factual errors.
\newline
4. Missing Necessary Information: The instruction lacks essential implementation details, required materials, data, formats, or context, making the task impossible to complete.
\newline
5. Safety, Privacy, or Confidentiality Risks: The instruction involves unsafe content, personal privacy leakage, company confidential information, or prohibited sensitive entities according to the annotation policy.
\newline
\textbf{Step 2: Reason Annotation} \newline
If you select "Low-quality" in Step 1, briefly explain the reason. The reason should identify the specific defect and be concise. Instructions judged as low-quality do not need further checklist revision.
\newline\newline
\textbf{[System Prompt]}\newline
\textcolor{red}{\{system\_prompt\}}\newline\newline
\textbf{[Multi-turn User Instruction]}\newline
\textcolor{red}{\{multi\_turn\_instruction\}}\newline\newline
Your judgment of instruction quality:
{\color{blue} \{option\}} A. High-quality \quad B. Low-quality \newline
Reason if B is selected: {\color{blue} \{reason\}}
\\
\bottomrule
\end{tabular}
}
\caption{Human annotation guideline for instruction quality inspection. The red part is the information provided to the annotators, and the blue part is the content that requires the annotators to make annotations. The repository policy file is provided as a system prompt to improve readability.}
\label{tab:instruction_quality_inspection_guideline}
\end{table*}

%% file: tables/checklist_revision_guideline.tex
\begin{table*}[t]
\small
\centering
\setlength{\tabcolsep}{1.6mm}{
\begin{tabular}{p{\dimexpr \linewidth-2\tabcolsep\relax}}
\toprule
Below, a high-quality multi-turn instruction, its corresponding constraint checklists at each turn, and its corresponding function checklist upon the conclusion of the multi-turn interaction will be provided.
Your task is to verify whether these checklists accurately cover the requirements of the instruction, and to modify them if any errors are found.
\newline
\newline
\textbf{Checklist Definition:} \newline
The constraint checklist contains the requirements that the model must follow when generating responses, including constraints from the system prompt and user instructions. 
The function checklist contains the functional requirements that should be satisfied upon the conclusion of the multi-turn interaction. 
Checklist items should be derived only from the provided instruction and should align with its requirements one-to-one.
\newline
\newline
\textbf{Task Details:}
\newline
\textbf{Step 1: Checklist Quality Judgment} \newline
Evaluate the correctness of both the constraint checklist and the function checklist. Select "Needs Revision" if any of the following conditions are satisfied:
\newline
1. Missing Core Requirements: The checklist omits mandatory constraints or core functional requirements explicitly required by the instruction.
\newline
2. Incorrect Description: A checklist item is not semantically equivalent to the original requirement, or it inaccurately paraphrases the instruction.
\newline
3. Fabricated Requirements: The checklist includes requirements that are not present in the system prompt or user instructions.
\newline
4. Obsolete Requirements: The checklist retains requirements that have been modified, replaced, or made inapplicable by later turns.
\newline
5. Unclear or Unverifiable Items: A checklist item is too vague, self-referential, contradictory, or not practically verifiable.
\newline
\textbf{Step 2: Checklist Modification} \newline
If you select "Needs Revision" in Step 1, revise the checklist by adding, modifying, or deleting items. You must follow these principles:
\newline
1. Source Integrity: Every checklist item must originate from the provided system prompt or user instructions.
\newline
2. Completeness and Accuracy: The revised checklist should cover all necessary constraints and core functional requirements without duplication or fabrication.
\newline
3. Minimal Sufficiency: Do not require the checklist to exhaustively cover every edge case, exceptional input, or minor visual detail unless it is explicitly required by the instruction.
\newline
4. Atomicity: Each checklist item should describe one verifiable requirement while remaining semantically complete.
\newline\newline
\textbf{[System Prompt]}\newline
\textcolor{red}{\{system\_prompt\}}\newline\newline
\textbf{[Multi-turn User Instruction]}\newline
\textcolor{red}{\{multi\_turn\_instruction\}}\newline\newline
\textbf{[Constraint Checklists at Each Turn]}\newline
\textcolor{red}{\{constraint\_checklists\}}\newline\newline
\textbf{[function checklist]}\newline
\textcolor{red}{\{functional\_checklist\}}\newline\newline
Your judgment of checklist quality:
{\color{blue} \{option\}} A. Correct \quad B. Needs Revision \newline
Main revision points if B is selected: {\color{blue} \{revision\_points\}}\newline
Your revised checklists if B is selected: {\color{blue} \{revised\_checklists\}}
\\
\bottomrule
\end{tabular}
}
\caption{Human annotation guideline for checklist revision. The red part is the information provided to the annotators, and the blue part is the content that requires the annotators to make annotations. The repository policy file is provided as a system prompt to improve readability.}
\label{tab:checklist_revision_guideline}
\end{table*}

%% file: tables/verification_code_generation_prompt.tex
\begin{table*}[!t]
\small
\centering
\setlength{\tabcolsep}{1.6mm}{
\begin{tabular}{p{\dimexpr \linewidth-2\tabcolsep\relax}}
\toprule
You are a code generation expert. 
I will provide you with a multi-turn user instruction, where the artificial intelligence assistant's responses in previous rounds are not given, as well as a specific constraint contained in the user instruction. 
My goal is to determine whether a certain artificial intelligence assistant's final-round response, together with the code project generated by that assistant, satisfies this constraint. 
Please generate a Python-format validation script for this constraint to help me make an accurate judgment. 
You must follow the principles below when completing this task:
\newline
\newline
1. The code must be a Python function. In this function:
\newline
\hspace*{1em} a. The function input parameters must include:
\newline
\hspace*{2em} (1) \texttt{response}: the content of the artificial intelligence assistant's final-round response.
\newline
\hspace*{2em} (2) \texttt{workspace\_path}: the path to the code project generated by the artificial intelligence assistant.
\newline
\hspace*{1em} b. The return value of the function must be a Boolean value, representing whether the constraint is satisfied.
\newline
\hspace*{1em} c. All packages that need to be imported in the function must be imported inside the function. Do not use any Python package that requires additional installation.
\newline
2. When counting words, splitting sentences, or counting emojis, in order to ensure a unified standard, you should directly use the following functions already implemented by human experts, unless the constraint gives special instructions, such as counting only English or Chinese characters. When using them, directly reference them in the code; do not reimplement them and do not import them:
\newline
\hspace*{1em} a. \texttt{count\_word(s: str) -> int}: counts the words of a string. The input parameter is a string, and the return value is an integer count. 
\newline
\hspace*{1em} b. \texttt{split\_sentences(s: str) -> List}: splits a string into sentences. The input parameter is a string, and the return value is a list containing each sentence in the string.
\newline
\hspace*{1em} c. \texttt{is\_emoji(s: str) -> bool}: determines whether a character is an emoji. The input parameter is the character, and the return value is a Boolean value indicating whether the character is an emoji.
\newline
3. Generate a validation script only for the given constraint. Do not generate the validation scripts for any other content in the user instruction.
\newline
\newline
The output format is as follows:
\newline
\newline
\textbf{[The Start of Analysis]}
\newline
... (You need to carefully analyze and think about how to generate a validation script to accurately judge whether the constraint is satisfied.)
\newline
\textbf{[The End of Analysis]}
\newline
\newline
\textbf{[The Start of Validation Script]}
\newline
```python
\newline
... (You need to provide Python-format code that can validate the constraint whether follow the specific constraints.)
\newline
```
\newline
\textbf{[The End of Validation Script]}
\newline
\newline
Below are the multi-turn user instruction I provide, and a specific constraint contained in the user instruction:
\newline
\textbf{[The Start of User Instruction]}
\newline
\textcolor{red}{\{{multi\_turn\_instruction\}}}
\newline
\textbf{[The End of User Instruction]}
\newline
\newline
\textbf{[The Start of Given Constraint]}
\newline
\textcolor{red}{\{constraint\}}
\newline
\textbf{[The End of Given Constraint]}
\\
\bottomrule
\end{tabular}
}
\caption{The prompt template for generating a validation script for code evaluation.}
\label{tab:validation_code_generation_prompt}
\end{table*}

%% file: tables/instruction_following_verification_prompt.tex
\begin{table*}[!t]
\small
\centering
\setlength{\tabcolsep}{1.6mm}{
\begin{tabular}{p{\dimexpr \linewidth-2\tabcolsep\relax}}
\toprule
You are an impartial judge, skilled at evaluating the quality of artificial intelligence assistants' responses and generated code. 
I will provide you with the operation trace of an artificial intelligence assistant responding to a user instruction, which may include the assistant's reasoning process, tool calls, tool-call results, and other information, the final response, a code project constructed based on the user instruction, and a list of constraints. 
Please think carefully and thoroughly analyze whether the assistant's operation trace, final response, and code project satisfy each constraint in the constraint list, and explain the reasons. 
You may use tools such as the terminal for evaluation. 
You must strictly follow the format below to output the analysis and judgment for each constraint:
\newline
\newline
```
\newline
\textbf{[The Start of Constraint 1]}
\newline
Constraint: ... (Directly provide the first constraint from the constraint list here, without making any modification.)
\newline
Analysis: ... (Based on the assistant's operation trace, final response, and the specific content of the code project, provide a detailed, step-by-step, and careful analysis of whether this constraint is satisfied.)
\newline
Conclusion: This can only be [[The constraint is satisfied]] or [[The constraint is not satisfied]].
\newline
\textbf{[The End of Constraint 1]}
\newline
\newline
\textbf{[The Start of Constraint 2]}
\newline
Constraint: ... (Directly provide the second constraint from the constraint list here, without making any modification.)
\newline
Analysis: ... (Based on the assistant's operation trace, final response, and the specific content of the code project, provide a detailed, step-by-step, and careful analysis of whether this constraint is satisfied.)
\newline
Conclusion: This can only be [[The constraint is satisfied]] or [[The constraint is not satisfied]].
\newline
\textbf{[The End of Constraint 2]}
\newline
\newline
...
\newline
```
\newline
\newline
\#\# Notes
\newline
1. The assistant's operation trace preserves the raw event stream of all operations performed by the assistant in this turn, except for the user instruction. The tool-call result will be replaced by a fixed placeholder, which is expected behavior.
\newline
2. The ``response'' only refers to the assistant's final response. The assistant's operation trace is not part of the response.
\newline
3. If a prerequisite basic requirement has problems and therefore makes the current constraint unverifiable, judge it as not satisfied.
\newline
4. Your judgment should be as strict as possible. Only when the assistant's operation trace, final response, and code project fully satisfy all parts of the corresponding constraint may you judge it as [[The constraint is satisfied]]. 
If there is any omission or error in satisfying the constraint, you should judge it as [[The constraint is not satisfied]].
\newline
5. Your judgments for the constraints in the constraint list should remain independent. 
When judging whether the current constraint is satisfied, you should not consider whether other constraints in the list are satisfied.
\newline
\newline
\#\# Assistant's Operation Trace
\newline
\textcolor{red}{\{trace\}}
\newline
\newline
\#\# Assistant's Final Response
\newline
\textcolor{red}{\{response\}}
\newline
\newline
\#\# Project Information
\newline
- Project description: \textcolor{red}{\{task\_description\}}
\newline
- Project source code: \textcolor{red}{\{workspace\_path\}}
\newline
\newline
\#\# Constraint List
\newline
\textcolor{red}{\{constraint\_checklist\}}
\\
\bottomrule
\end{tabular}
}
\caption{The prompt template for verifying the constraint checklist item.}
\label{tab:instruction_following_verification_prompt}
\end{table*}

%% file: tables/function_verification_prompt_http.tex
\begin{table*}[!t]
\small
\centering
\setlength{\tabcolsep}{1.6mm}{
\begin{tabular}{p{\dimexpr \linewidth-2\tabcolsep\relax}}
\toprule
You are a software testing expert. You need to test a software project according to a specific checklist item and determine whether the current software project satisfies the requirement of that checklist item. You may use various tools, including Playwright and the terminal, and make use of your visual understanding and code testing capabilities to evaluate the software project efficiently and accurately.
\newline
\newline
\#\# Notes
\newline
1. The project has already been built and deployed to a local HTTP service, and you can access it directly through the browser.
\newline
2. Before the evaluation process, you need to complete the necessary preparation work:
\newline
\hspace*{1em} a. Read the software project's README file, if it exists, and follow its instructions to configure the environment and use the relevant features.
\newline
\hspace*{1em} b. If there is no README, configure the environment according to the needs of the project. Do not fix environment problems that already exist in the project code itself, such as incorrect library import paths; such cases should be regarded as not satisfying the checklist item.
\newline
3. During the evaluation process, you must strictly follow the way a real user would use the project.
\newline
4. When using Playwright for testing, you may only use keyboard and mouse operations that a user can perform, such as clicking and typing. You must not use evaluation methods that are inconsistent with ordinary user behavior.
\newline
5. If a prerequisite feature in the software has problems, causing the feature corresponding to the current checklist item to be untestable, meaning the user cannot reach that feature, then the current checklist item should be regarded as not satisfied.
\newline
6. When making the judgment, you must strictly evaluate from the user's perspective.
\newline
7. If a feature is correctly designed at the code level but has obvious problems in the UI, the feature should be regarded as not satisfying the requirement.
\newline
8. When taking screenshots with Playwright, important information must be included within the screenshot frame, such as by using a full-page screenshot. A judgment may only be made when the screenshot information is complete.
\newline
\newline
\#\# Project Information
\newline
- Project description: \textcolor{red}{\{task\_description\}}
\newline
- Project access URL: \textcolor{red}{\{project\_url\}}
\newline
- Project source code: \textcolor{red}{\{workspace\_path\}}
\newline
\newline
\#\# Current Checklist Item
\newline
\textcolor{red}{\{checklist\_item\}}
\newline
\newline
\#\# Output Format
\newline
If the project does not satisfy the \textbf{current checklist item}, please output ``Judgment Conclusion: The project does not satisfy the requirement''. If the project satisfies the current checklist item, please output ``Judgment Conclusion: The project satisfies the requirement''.
\newline
\newline
Please start your evaluation.
\\
\bottomrule
\end{tabular}
}
\caption{The prompt template for verifying the function checklist item for frontend development task.}
\label{tab:function_checklist_verification_http}
\end{table*}

%% file: tables/function_verification_prompt_other.tex
\begin{table*}[!t]
\small
\centering
\setlength{\tabcolsep}{1.6mm}{
\begin{tabular}{p{\dimexpr \linewidth-2\tabcolsep\relax}}
\toprule
You are a software testing expert. You need to test a software project according to a specific checklist item and determine whether the current software project satisfies the requirement of that checklist item. 
You may use various tools, including Playwright and the terminal, and make use of your visual understanding and code testing capabilities to evaluate the software project efficiently and accurately.
\newline
\newline
\#\# Notes
\newline
1. Before the evaluation process, you need to complete the necessary preparation work:
\newline
\hspace*{1em} a. Read the software project's README file, if it exists, and follow its instructions to configure the environment and use the relevant features.
\newline
\hspace*{1em} b. If there is no README, configure the environment according to the needs of the project. Do not fix environment problems that already exist in the project code itself, such as incorrect library import paths; such cases should be regarded as not satisfying the checklist item.
\newline
2. During the evaluation process, you must strictly follow the way a real user would use the project.
\newline
3. When using Playwright for testing, you may only use keyboard and mouse operations that a user can perform, such as clicking and typing. You must not use evaluation methods that are inconsistent with ordinary user behavior.
\newline
4. If a prerequisite feature in the software has problems, causing the feature corresponding to the current checklist item to be untestable, meaning the user cannot reach that feature, then the current checklist item should be regarded as not satisfied.
5. When making the judgment, you must strictly evaluate from the user's perspective.
\newline
6. If a feature is correctly designed at the code level but has obvious problems in the UI, the feature should be regarded as not satisfying the requirement.
\newline
7. When taking screenshots with Playwright, important information must be included within the screenshot frame, such as by using a full-page screenshot. A judgment may only be made when the screenshot information is complete.
\newline
\newline
\#\# Project Information
\newline
- Project description: \textcolor{red}{\{task\_description\}}
\newline
- Project source code: \textcolor{red}{\{workspace\_path\}}
\newline
\newline
\#\# Current Checklist Item
\newline
\textcolor{red}{\{checklist\_item\}}
\newline
\newline
\#\# Output Format
\newline
If the project does not satisfy the \textbf{current checklist item}, please output ``Judgment Conclusion: The project does not satisfy the requirement''. If the project satisfies the current checklist item, please output ``Judgment Conclusion: The project satisfies the requirement''.
\newline
\newline
Please start your evaluation.
\\
\bottomrule
\end{tabular}
}
\caption{The prompt template for verifying the function checklist item for other tasks.}
\label{tab:function_checklist_verification_non_http}
\end{table*}